\documentclass[letterpaper]{article} 
\usepackage{aaai2026}  
\usepackage{times}  
\usepackage{helvet}  
\usepackage{courier}  
\usepackage[hyphens]{url}  
\usepackage{graphicx} 
\usepackage{natbib}  
\usepackage{caption} 
\usepackage{algorithm}
\usepackage{algorithmic}

\usepackage{amsmath}
\usepackage{amsfonts,amssymb}
\usepackage{xcolor}

\usepackage{booktabs} 
\usepackage{subfigure}

\newtheorem{theorem}{Theorem}

\newtheorem{corollary}[theorem]{Corollary}

\usepackage{newfloat}
\usepackage{listings}
\DeclareCaptionStyle{ruled}{labelfont=normalfont,labelsep=colon,strut=off} 
\floatstyle{ruled}
\newfloat{listing}{tb}{lst}{}
\floatname{listing}{Listing}
\title{Group Causal Policy Optimization for Post-Training Large Language Models}
\author{
    Ziyin Gu\textsuperscript{\rm 1}\textsuperscript{\rm 2}\equalcontrib,
    Jingyao Wang\textsuperscript{\rm 1}\textsuperscript{\rm 2}\equalcontrib,
    Ran Zuo\textsuperscript{\rm 3},
    Chuxiong Sun\textsuperscript{\rm 1}\textsuperscript{\rm 2},
    Zeen Song\textsuperscript{\rm 1}\textsuperscript{\rm 2},
    Changwen Zheng\textsuperscript{\rm 1}\textsuperscript{\rm 2},
    Wenwen~Qiang\textsuperscript{\rm 1}\textsuperscript{\rm 2}\thanks{Corresponding author. Email: qiang.ww0922@gmail.com}
}
\affiliations{
    \textsuperscript{\rm 1}Institute of Software, Chinese Academy of Sciences, Beijing, China\\
    \textsuperscript{\rm 2}University of the Chinese Academy of Sciences, Beijing, China\\
    \textsuperscript{\rm 3}Communication University of China, Beijing, China

}

\usepackage{bibentry}

\begin{document}

\maketitle

\begin{abstract}
Recent advances in large language models (LLMs) have broadened their applicability across diverse tasks, yet specialized domains still require targeted post-training. Among existing methods, Group Relative Policy Optimization (GRPO) stands out for its efficiency, leveraging groupwise relative rewards while avoiding costly value function learning. However, GRPO treats candidate responses as independent, overlooking semantic interactions such as complementarity and contradiction. To address this challenge, we first introduce a Structural Causal Model (SCM) that reveals hidden dependencies among candidate responses induced by conditioning on a final integrated output—forming a collider structure. Then, our causal analysis leads to two insights: (1) projecting responses onto a causally-informed subspace improves prediction quality, and (2) this projection yields a better baseline than query-only conditioning. Building on these insights, we propose Group Causal Policy Optimization (GCPO), which integrates causal structure into optimization through two key components: a causally-informed reward adjustment and a novel KL-regularization term that aligns the policy with a causally-projected reference distribution. Comprehensive experimental evaluations demonstrate that GCPO consistently surpasses existing methods—including GRPO—across multiple reasoning benchmarks.
\end{abstract}


\section{Introduction}

Recent advances in large language models (LLMs) have significantly broadened their application potential, demonstrating remarkable capabilities in general tasks \cite{lai2025survey, zhao2023survey, minaee2024large, jaech2024openai}. However, fully harnessing their practical effectiveness, particularly in specialized domains, requires focused post-training adjustments \cite{tie2025survey}. While foundational pre-training establishes linguistic fluency and general reasoning, supplementary methods such as reinforcement learning with human feedback (RLHF) \cite{bai2022training} are essential for adapting LLMs to specific applications and aligning their outputs with human preferences and ethical norms. Among these approaches, the recently proposed Group Relative Policy Optimization (GRPO) \cite{shao2024deepseekmath} has garnered considerable attention due to its significant reduction in computational overhead and memory requirements. By introducing a scalable and efficient training mechanism, GRPO has demonstrated substantial performance gains on many benchmarks \cite{guo2025deepseek}.

\begin{figure}[t]
    \centering
    \includegraphics[width=\linewidth]{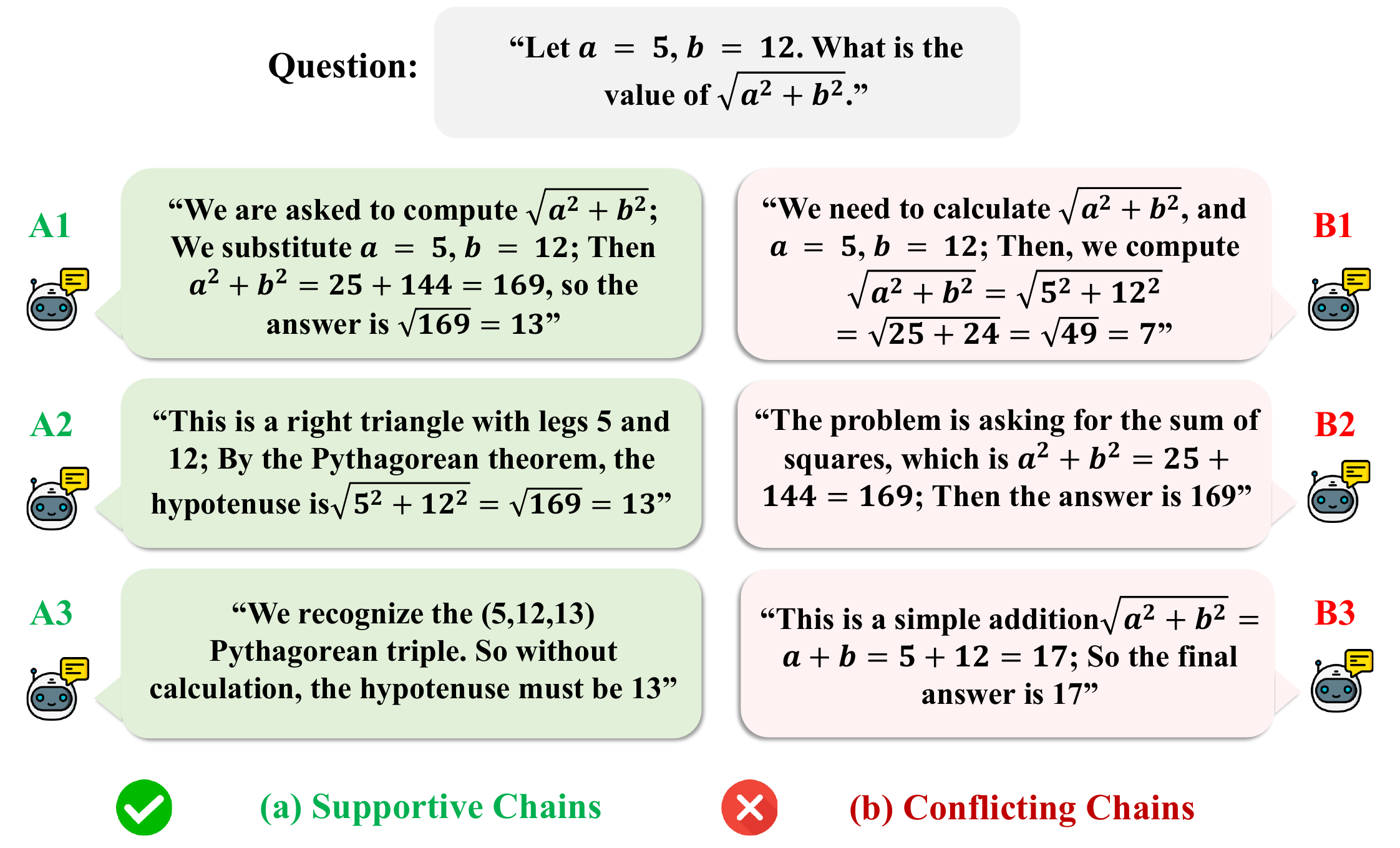}
    \caption{Examples of (a) Supportive chains: A1 provides precise computation, A2 offers geometric insight, and A3 quickly verifies the result via Pythagorean triple recognition; combined, they robustly lead to the optimal answer 13; (b) Conflicting chains: B1 yields 7 due to calculation errors, B2 outputs 169 by omitting the square root, and B3 misinterprets the question to give 17; their conclusions conflict, and mixing them with correct paths introduces errors.}
    \label{exapasd}
\end{figure}

While GRPO introduces an efficient mechanism by estimating advantages through groupwise relative rewards, it adopts a simplifying assumption: all candidate responses within a group are treated as independent and unrelated. This design choice helps reduce computational complexity and makes the method more scalable, especially when compared to traditional value-based approaches like PPO \cite{schulman2017proximal,ouyang2022training}. However, in many real-world reasoning tasks, responses generated for the same input often contain rich semantic connections. For an example shown in Figure \ref{exapasd}, some responses may complement each other by covering different aspects of the problem, jointly forming a more complete reasoning chain; others may contradict each other, revealing logical conflicts or alternative interpretations. These interactions—whether supportive or conflicting—are not captured in the current formulation of GRPO. As a result, although GRPO successfully leverages relative reward signals within a group, it may overlook valuable information encoded in the relationships between responses. Incorporating such intra-group dynamics could enable models to better understand the structure of the output space, leading to more nuanced learning and potentially improved alignment with human reasoning preferences.

To address this challenge from a causal perspective, we introduce a Structural Causal Model (SCM) that explicitly captures the relationships between the original query and the generated candidate responses. Specifically, consider a scenario in Figure~\ref{fig_SCM} where a user inputs a query into the LLM, resulting in multiple independently generated candidate answers. Initially, these candidate outputs seem unrelated since each is generated independently based solely on the query. However, if we subsequently use these candidate answers collectively to produce a final, refined response, we unintentionally create a collider structure. In causal inference terms \cite{pearl2016causal, pearl2009causality}, a collider is a scenario where two or more independent variables influence a common variable, such that conditioning on this common variable makes these previously independent variables become interdependent. In our case, candidate responses are initially independent when conditioned solely on the query. But when these responses jointly influence a final integrated output, conditioning on this final result (the collider) introduces dependencies among the candidate responses. Practically speaking, knowing the content of the final integrated response can reveal previously hidden relationships among candidate answers. For example, one candidate response might provide context missing from another, forming a complementary relationship; another might present contradictory logic, creating a conflicting relationship. Recognizing and explicitly modeling these collider-induced relationships might help the model better leverage hidden structural patterns within generated answers.

Formally, our causal analysis (refer to Section: Causal Analysis and Motivation for more details) provides a rigorous theoretical basis for this intuition. Specifically, Theorem~\ref{pro_dsf} indicates that when the query-response generation process follows a collider structure, predicting an output based on a causally adjusted baseline—that is, the projection of the original predictions onto a subspace that respects this collider structure—will consistently yield improved accuracy. In other words, rather than directly predicting responses based solely on independent evaluations, incorporating a causally informed adjustment significantly enhances prediction performance. Moreover, Corollary~\ref{cor_dswerf} complements this by showing that even the original query-based predictions can benefit from incorporating this causally projected baseline. Intuitively, this can be thought of as adding a causal ``lens'' through which predictions are viewed, enabling the model to correct latent biases or misunderstandings that arise from ignoring structural dependencies.

Motivated by these causal insights, we propose a novel optimization method called Group Causal Policy Optimization (GCPO). Unlike GRPO, which evaluates each candidate response purely based on its reward relative to the group average, GCPO explicitly incorporates causal relationships within the group of generated outputs. Guided by Theorem~\ref{pro_dsf} and Corollary~\ref{cor_dswerf}, GCPO introduces two major adjustments to the original GRPO framework: (1) a causally-adjusted reward mechanism, and (2) a novel KL-divergence regularization term that explicitly considers causal structures. First, the reward mechanism in GCPO is enhanced by projecting each candidate response onto a causally-informed baseline. Practically, the reward of each candidate answer is adjusted based on how closely it aligns with this causally projected reference. Intuitively, this approach rewards responses that are not only individually strong but also structurally coherent with other responses. Second, to further encourage structural consistency, GCPO introduces an additional KL-divergence regularization term. Specifically, during training, we first compute the model's output distribution conditioned solely on the query. Next, we calculate a causally-adjusted distribution that captures interdependencies among candidate responses. The sum of these two components forms a new reference distribution, representing the model’s corrected belief after considering group-level causal structures. By minimizing the KL-divergence between the model’s current output and this causally-informed reference, GCPO explicitly guides the model towards structurally consistent predictions. To further illustrate intuitively, the original GRPO method measures divergence by comparing the current policy model to a standard reference model trained without causal adjustments. GCPO, however, measures this divergence against a structurally enhanced baseline, explicitly encouraging the policy model to conform to inferred dependencies among candidate responses. The main contributions of this paper can be summarized as:
\begin{itemize}
    \item Causal insight into candidate dependencies. We establish that conditioning on a final integrated output induces a collider structure among candidate responses. Theoretically, we prove that projecting predictions onto a causally-informed subspace reduces test error, offering a more reliable baseline than query-only conditioning.
    \item A causality-aware policy optimization method. We propose GCPO, which enhances GRPO with a causally-adjusted reward and a KL regularizer aligned to a projected reference distribution. This enables structurally consistent and semantically robust policy updates.
    \item Consistent gains across benchmarks. Experiments on math and code reasoning tasks show that GCPO consistently outperforms GRPO. Ablations confirm the critical role of both proposed components.
\end{itemize}

\section{Related Work}
In recent years, LLMs have made remarkable progress on a wide range of tasks, including question answering \cite{bottou2018optimizationmethodslargescalemachine,bai2024digirltraininginthewilddevicecontrol}, code generation \cite{sadik2025benchmarkingllmcodesmells,wang2025insightsverificationtrainingverilog}, and mathematical reasoning \cite{minaee2024survey,wang2025learning,muennighoff2025s1}.
However, achieving optimal performance on specialized tasks often requires targeted post-training adaptation \cite{tie2025survey}.
Common approaches such as Supervised Fine-Tuning \cite{raffel2020exploring,devlin2019bert} and Instruction Tuning \cite{sanh2022multitask,chung2022scaling,ouyang2022training} use labeled data or instructional examples to align model outputs with specific objectives, delivering strong results.
Nevertheless, these post-training methods are prone to exposure bias and may generalize poorly to novel scenarios \cite{touvron2023llama,ballon2025relationship}.

To address these limitations, reinforcement learning (RL) has been adopted to tailor LLMs for domain-specific applications and align their outputs with human preferences and ethical standards \cite{ouyang2022training,bai2022training}.
Under RL-based strategies, GRPO \cite{shao2024deepseekmath} has garnered widespread attention with its efficiency in lowering computational and memory burdens. It introduces a scalable group-wise optimization framework, where policy updates leverage relative advantages within groups of candidate responses.
This design enables flexible integration of process rewards and preference signals, resulting in great performance.
Building upon GRPO, a number of variants have been proposed, leveraging process-level reward estimation, adaptive reward shaping, and regularization strategies to further improve efficiency and generalization. Specifically, LC-R1 \cite{cheng2025optimizing} employs a novel
combination of a length reward for overall conciseness and a compress reward that is specifically designed to remove the invalid portion of the thinking process. GVPO \cite{zhang2025gvpo} incorporates the analytical solution to KL-constrained reward maximization directly into its gradient weights, ensuring alignment with the optimal policy. Dr.GRPO \cite{liu2025understanding} improves token efficiency while maintaining reasoning performance. L2T \cite{wang2025learning} proposes an information-theoretic reinforcement fine-tuning framework for LLMs to make the models achieve optimal reasoning with fewer tokens. 

However, these exist RL-based policy optimization  methods often treat candidate responses as independent, thus ignoring the rich structural and causal relationships that are embedded in the interrelationships among responses. To address this, in this work, we propose GCPO that explicitly models and leverages intra-group dependencies to improve the general coherence and reasoning capability of LLMs.

\section{Causal Analysis and Motivation} \label{qww_1}

This section begins by introducing an SCM. Based on this foundation, we construct a causal analysis framework to evaluate the quality of reasoning strategies in LLMs. We conclude by outlining the motivation that informs the design of the proposed approach.

\subsection{Causal Analysis}
\begin{figure}[t]
\centering
\includegraphics[width=0.6\columnwidth]{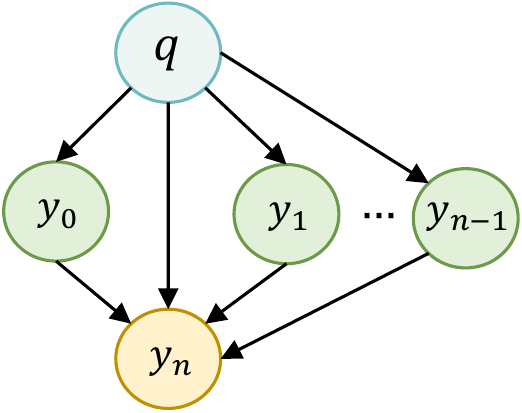}
\caption{The SCM under our setting. $q$ is the input query, $\{y_0,\cdots,y_{n-1}\}$ represents the set of outputs obtained by feeding $q$ into a LLM $n$ times, and $y_n$ denotes the final output produced by inputting $\{q,y_0,\cdots,y_{n-1}\}$ into a LLM.}
\label{fig_SCM}
\end{figure}

Consider an SCM illustrated in Figure~\ref{fig_SCM}. Here, the variable $q$ represents the original input query. The variables $y_0, y_1, \cdots, y_{n-1}$ respectively denote the corresponding outputs obtained by independently feeding the same query $q$ into the function $\pi$. The variable $y_n$ is a new output derived by feeding $q, y_0, y_1,\cdots,y_{n-1}$ into the $\pi$. Consequently, the SCM includes causal paths: $\{q \to y_i \to y_n\}_{i=0}^{n-1}$ and $q \to y_n$. In addition, the path $q \to \{y_0,y_1,\cdots, y_{n-1}\}$ forms a fork structure, while the path $\{y_0,y_1,\cdots, y_{n-1}\} \to y_{n}$ forms a collider structure \cite{pearl2009causality}. These structures lead to two conditional independence relations \cite{pearl2016causal}: conditioned on $q$, the variables in $\{y_i\}_{i=0}^{n-1}$ are mutually independent; however, conditioned additionally on $y_n$, these variables become mutually dependent.

From a Bayesian perspective, when a model is trained to optimality by minimizing the cross-entropy loss or mean squared error loss, it can be viewed as estimating the conditional expectation of the output distribution given the input context \cite{zhang2018language, goodfellow2016deep, bengio2003neural}. More precisely, if the function $\pi$ is trained using cross-entropy loss and achieves optimality, then statistical decision theory implies that $\pi(X) = \mathbb{E}[Y \mid X]$, where $X$ refers to the input in a general sense, and $Y$ refers to the corresponding output. For clarity of distinction, let $\pi$ denote a function that outputs a probability, similar to the formulation used during training based on cross-entropy loss minimization, and let $\pi^*$ denote its Bayesian optimal counterpart. Then, we have $\pi(y_0 \mid q) = p_{\pi}(y_0\mid q)$ and $\pi^*(q) = \mathbb{E}[y_0 \mid q]$. Based on the conditional independence relations discussed in the previous paragraph, the following conclusion can be drawn:
\begin{equation}\label{eq:esp}
    \begin{array}{l} 
      \quad  \mathbb{E}\left [ \pi^*(x) - \pi^*(q) \mid q,  y_{1:n-1} \right ] \\
=\mathbb{E}\left [ \mathbb{E}\left [ y_0\mid x \right ] - \mathbb{E}\left [ y_0\mid q  \right ] | q, y_{1:n-1} \right ] \\ 
=\mathbb{E}\left [ y_0\mid q, y_{1:n-1} \right ] -  \mathbb{E}\left [ y_0 \mid q, y_{1:n-1} \right ] =0,\\ 
\end{array} 
\end{equation}
where $x = \{q, y_1, \cdots, y_{n}\}$, $y_{1:n-1}=\{y_1, \cdots, y_{n-1}\}$. It is important to note that the variables $y_0, y_1, \cdots, y_{n-1}$ are used as generic placeholders. In other words, Equation (\ref{eq:esp}) still holds when $y_0$ is exchanged with any $y_i \in \{y_1, \cdots, y_{n-1}\}$. All subsequent results in this section follow this property, and we will not reiterate it in the follows.

Let $\mathcal{F}$ denote the space of square-integrable functions, we can obtain that $\pi^* \in \mathcal{F}$. Let $\Phi$ be a functional operator acting on $\pi^*(x)$, defined as the following:
\begin{equation} \label{eq:dfh}
\Phi\cdot\pi^*(x)=\mathbb{E}[\pi^*(x)|q, y_1,\cdots,y_{n-1}].
\end{equation}
Based on this, we define a causal-related mapping $\Psi$ as: $\Psi = {\rm Id} - \Phi$, where ${\rm Id}$ denotes the identity mapping. Noting that the output of $\Phi$ can be viewed as the image space, while the output of $\Psi$ corresponds to the kernel space associated with $\Phi$. Meanwhile, let $X$ be the random variable of the query and $Y$ be the random variable of the answer, assume $X \times Y \sim p(X,Y)$ where $p(X,Y)$ is the joint probability distribution of $X$ and $Y$. Given $\pi^*_1, \pi^*_2 \in \mathcal{F} $, $\Delta$ is defined as:
    \begin{equation}\label{eq_aagsd}
\begin{array}{l} 
    \Delta (\pi^*_1,\pi^*_2) =  \underset{ p(X,Y)}{\mathbb{E}}  \left [ Y - \pi^*_1(X) \right ]\\
\quad\quad\quad\quad\quad\quad\quad\quad\quad -\underset{p(X,Y)}{\mathbb{E}}  \left [ Y - \pi^*_2(X) \right ]. 
\end{array} 
\end{equation}
Equation (\ref{eq_aagsd}) can be interpreted as the test error or the expected risk. Then, the following conclusion can be drawn:
\begin{theorem}\label{pro_dsf} 
    Given the condition of Equation (\ref{eq:esp}) and  the SCM shown in Figure~\ref{fig_SCM}, for $\forall \pi^* \in \mathcal{F} $, the following holds:
\begin{equation}\label{eq_asgfh}
\begin{array}{l} 
    \Delta (\pi^*(x),\Psi \cdot \pi^*(x)+\pi^*(q)) \ge 0.
\end{array} 
\end{equation}   
\end{theorem}

The proof of Theorem \ref{pro_dsf} is presented in the Appendix. We provide an intuitive understanding of Theorem \ref{pro_dsf}. First, the collider structure makes us realize that, although some variables may appear independent on the surface, they could potentially be dependent through a common influence. If this relationship is not captured, it may affect the accuracy of the LLM. Second, $\pi^*(x)$ is tasked with predicting an outcome based on the input. From a causal perspective, $\pi^*(x)$ serves as a generative function. If we know that the data generation process follows a collider structure, we can project the hypothesis space formed by all $\pi^*(x)$ onto a subspace formed by those $\pi^*(x)$ that can recognize the collider structure. This is akin to adding a pair of ``glasses'' to $\pi^*(x)$, helping it identify latent dependencies that are not immediately apparent, thereby improving its predictive accuracy on new data. Furthermore, $\pi^*(q)$ represents the model's initial prediction in the absence of the collider structure’s influence, and it can be viewed as a preliminary estimate of the input. By incorporating this initial estimate, we can further optimize the model, ensuring that the final output does not merely rely on the preliminary estimate but fully considers the inherent structure of the data. Similarly, the follows can also be drawn:

\begin{corollary}\label{cor_dswerf} 
    Given the condition of Equation (\ref{eq:esp}) and  the SCM shown in Figure~\ref{fig_SCM}, for $\pi^*$ and $\Psi$, the following holds:
\begin{equation}\label{eq_asfdh}
    \Delta (\pi^*(q),\Psi \cdot \pi^*(x)+\pi^*(q)) \ge 0.
\end{equation}   
\end{corollary}

The proof of Corollary \ref{cor_dswerf} is provided in the Appendix. Since the intuitive interpretation of Corollary \ref{cor_dswerf} closely parallels that of Theorem \ref{pro_dsf}, we omit a redundant explanation. Together, Theorem \ref{pro_dsf} and Corollary \ref{cor_dswerf} suggest that when the query generation process involves a collider structure, it is possible to project the hypothesis space of an LLM onto a subspace that better aligns with this structure. By incorporating a baseline function, the model can be further optimized. This approach leverages conditional independence relations encoded in the causal graph, thereby improving the generalization capability of the LLM and enabling more stable and reliable performance on unseen queries.

\subsection{Motivation Analysis}
GRPO has been widely adopted for post-training LLMs due to its efficiency and simplicity. It treats the model as a policy and optimizes it by comparing relative rewards among candidate responses generated for the same query. However, GRPO assumes that all candidates are independent, overlooking potential semantic interactions such as complementarity or contradiction (see Figure~\ref{exapasd}). This limits the reward signal expressiveness and may hinder LLMs generalization.

From the above causal analysis, while candidate responses are independently sampled from the query, they often influence a final integrated output, thus forming a collider structure. Under this structure, responses become conditionally dependent when the final output is observed. Our theoretical findings (Theorem~\ref{pro_dsf} and Corollary~\ref{cor_dswerf}) show that projecting predictions onto a causally-informed subspace, expressed as $\Psi \cdot \pi^*(x) + \pi^*(q)$, indeed leads to consistently lower test error than using $\pi^*(q)$ or $\pi^*(x)$ alone.

This insight motivates a principled revision of GRPO’s preference mechanism. Instead of favoring candidates purely based on relative rewards, we can additionally consider their alignment with the causally projected output. This adjustment allows the model to exploit latent dependencies among responses, encouraging structurally coherent and semantically accurate outputs. Furthermore, we can introduce a causal regularization term that aligns the policy with a causally-informed reference distribution. Together, these changes form the basis of the following proposed GCPO, a causality-aware optimization framework that enhances model performance by integrating structural reasoning signals into the learning process.

\section{The Proposed Method}

In this section, we propose GCPO, a new post-training algorithm for LLMs. GCPO can be viewed as a variant of GRPO, with the primary differences lying in two aspects: the relative advantage function and the KL Divergence.

\subsection{A Brief Introduction to GRPO}

For a given query $q$, GRPO samples a set of outputs $\{y_0, y_1, \cdots, y_{n-1}\}$ from the old policy $\pi_{\theta_{\rm old}}$. It then updates the policy $\pi_{\theta}$ by maximizing the objective:
\begin{equation}\label{eq_grpo}
    \begin{array}{c} 
\mathcal{J}_{\rm GRPO} = \mathbb{E}_{[q \sim P,\{y_0,y_1,\cdots,y_{n-1}\} ]} \\ 
  \frac{1}{n} \sum\limits_{i=0}^{n-1} \frac{1}{T_i} \sum\limits_{j=0}^{T_i} \{ [\min(R_{i,j}(\theta)A_{i}, \Xi_{ij} \cdot A_i)] \\
  -\beta {{ D}_{\rm KL}} (\pi_{\theta}\parallel \pi_{\rm ref}) \},
\end{array} 
\end{equation} 
where $\epsilon$ and $\beta$ are hyperparameters, $\Xi_{ij} = {\rm clip}(R_{i,j}(\theta),1-\epsilon ,1+\epsilon )$, $\mathrm{clip}(\cdot)$ is a truncation function ensuring stable updates, $P$ denotes the distribution of queries. Because an LLM generates a output $y_i = (y_{i,1}, \dots, y_{i,T_i})$ token‑by‑token in an autoregressive manner, where $T_i$ denotes the token length of $y_i$, thus, $R_{i,j}(\theta)$ and ${D}_{\rm KL}(\cdot)$ are also calculated in a token‑by‑token manner. Then, the KL divergence term ${D}_{\rm KL}(\pi_{\theta} \parallel \pi_{\rm ref})$ is computed as:
\begin{equation}\label{eq_gfdgrpo}
\frac{\pi_{\rm ref}(y_{i,j}|q,y_{i,<j})}{\pi_{\theta}(y_{i,j}|q,y_{i,<j})} -\log \frac{\pi_{\rm ref}(y_{i,j}|q,y_{i,<j})}{\pi_{\theta}(y_{i,j}|q,y_{i,<j})}-1,
\end{equation}
where $\pi_{\rm ref}$ is a fixed reference policy and often set to $\pi_{\theta_{\rm old}}$. The relative advantage $A_i$ is calculated within each sampled group to capture the comparative quality of outputs:
\begin{equation}\label{eq_gfqwegrpo}
A_i=[r_i-{\rm mean}(r_0,\cdots,r_{n-1})]/{\rm std}(r_0,\cdots,r_{n-1}), 
\end{equation}
where $r_i = \mathrm{reward}(y_i)$ combines task-specific accuracy and formatting rewards, while $\mathrm{mean}(\cdot)$ and $\mathrm{std}(\cdot)$ are the mean and standard deviation over the reward group.
At last, the importance ratio $R_{i,j}(\theta)$ is defined as:
\begin{equation}\label{eq_gsdfo}
\pi_{\theta}(y_{i,j}|q,y_{i,<j}) / \pi_{\theta_{\rm old}}(y_{i,j}|q,y_{i,<j}).
\end{equation}

\subsection{Details of the Proposed GCPO}
For a query $q$, GCPO also samples a group of outputs $\{y_0,y_1,\cdots,y_{n-1}\}$ from the old policy $\pi_{\theta_{\rm old}}$. Different from GRPO, GCPO then input $q$ and $\{y_0,y_1,\cdots,y_{n-1}\}$ into the old policy $\pi_{\theta_{\rm old}}$ for $n$ times to obtain a final outputs $y_n$ and $\{y_{n,i}\}_{i=1}^{n-1}$. GCPO optimizes the policy model $\pi_{\theta}$ by maximizing the following objective:
\begin{equation}\label{eq_gcpo}
\begin{array}{c} 
\mathcal{J}_{\rm GCPO} = \mathbb{E}_{[q\sim P,\{y_0,y_1,\cdots,y_n\},\{y_{n,i}\}_{i=1}^{n-1} ]} \\ 
\frac{1}{n} \sum\limits_{i=0}^{n-1} \frac{1}{T_i} \sum\limits_{j=0}^{T_i} \{ [\min(R_{i,j}(\theta)B_i,\Xi_{ij} \cdot B_i)] \\
\ -\beta {D_{\rm KL}} (\pi_{\theta}\parallel \pi_{\rm ref}) \}- \kappa {D_{\rm KL}} (\pi_{\theta}\parallel \pi'_{\rm ref}) ,
\end{array} 
\end{equation}
where $\kappa$ is a hyper-parameter, $B_i$ is the newly proposed relative advantage function, $\pi'_{\rm ref}$ is the newly proposed pre-defined policy model, and ${D_{\rm KL}} (\pi_{\theta} \parallel \pi'_{\rm ref})$ is the newly proposed regularizer. In the following, we conduct an in-depth study of the terms $B_i$ and ${D_{\rm KL}} (\pi_{\theta} \parallel \pi'_{\rm ref})$. 


\textbf{Design of Relative Advantage Function}. Formally, we define $B_i = A_i \cdot \Upsilon_i$, where $A_i$ is computed in the same way as in GRPO. We next describe the procedure for designing and computing $\Upsilon_i$. The design of $\Upsilon_i$ is inspired by Corollary~\ref{cor_dswerf}. According to Equation~(\ref{cor_dswerf}), when we focus on the answer variable $y_0$, the expected risk of the output from $\pi^*(q)$ is higher than that of $\Psi \cdot \pi^*(x) + \pi^*(q)$. This suggests, conservatively, that the output quality of $\Psi \cdot \pi^*(x) + \pi^*(q)$ is better than that of $\pi^*(q)$. Based on this observation, the advantage value for each candidate answer corresponding to a given query can be designed as follows: the closer the candidate is to the output of $\Psi \cdot \pi^*(x) + \pi^*(q)$, the higher the advantage value it receives.

The proposed approach faces two practical challenges during implementation: (1) how to approximate $\Psi \cdot \pi^*(x) + \pi^*(q)$; and (2) how to measure the similarity between model outputs. Since $\pi^*$ represents concrete generated content, which typically includes both intermediate reasoning steps and the final answer, we propose to approximate $\Psi \cdot \pi^*(x) + \pi^*(q)$ using the feature representation of the output. Then, we measure similarity based on the cosine distance between these feature representations. The detailed procedure is as follows: \textbf{ Step 1:} Approximating $\pi^*_{\theta}(q)$.
Given a answer $y_i$, we define $o_i$ as the combination of $y_i$ and the intermediate reasoning steps leading to it. We then feed $o_i$ back into $\pi_{\theta}$, and extract the hidden representation of the final token from the last layer as the feature representation of $o_i$, denoted by $z_i$. Then, let $\bar{z} = {\rm mean} (z_0,\cdots, z_{n-1})$, which can be regarded as a Monte Carlo approximation of the output representation of $\pi^*_{\theta}(q)$;
\textbf{Step 2:} Approximating $\pi^*_{\theta}(x)$ for $y_i \in \{y_i\}_{i=0}^{n-1}$. Because that $\pi^*_{\theta}(x) = \mathbb{E}\left [ y_0 \mid x \right ]$, when we focus on $y_i$, it equals to that $y_0$ is exchanged with $y_i$, and the condition $x$ is exchanged with $x_i = \{q, y_0, \cdots, y_{n}\} \setminus \{y_i\}$. We then feed $x_i$ into $\pi_{\theta}$ for $n$ times to obtain the corresponding outputs $\{O_{i,j}\}_{j=1}^{n}$ and representations $\{Z_{i,j}\}_{j=1}^{n}$. Let $\bar{Z}_i = {\rm mean} (Z_{i,1},\cdots, Z_{i,n})$, which can be regarded as a Monte Carlo approximation of the output representation of $\pi^*_{\theta}(x)$ for $y_i$;
\textbf{Step 3:} Approximating $\Phi \cdot \pi^*(x)$ for $y_i \in \{y_i\}_{i=0}^{n-1}$. According to the definition of $\Phi$, we first define $y_{n,0}=y_n$ and $x_{i,j} = \{q, y_0, \cdots, y_{n,j}\} \setminus \{y_i\}$, where $j \in \{0,\cdots,n-1\}$.
We repeat ``Step 2'' for the set $\{x_{i,j}\}_{j=0}^{n-1}$ to obtain the corresponding representations $\{\bar{Z}_{i,j}\}_{j=0}^{n-1}$. The average of it is denoted by $\bar{Z'}_i$, which serves as a Monte Carlo approximation of the output representation of $\Phi \cdot \pi^*(x)$ for $y_i$.
\textbf{Step 4:} Approximating $\Psi \cdot \pi^*(x) + \pi^*(q)$.
Combining the previous steps, $\bar{Z}_i - \bar{Z'}_i + \bar{z}$ serves as a Monte Carlo approximation of the output representation of $\Psi \cdot \pi^*(x) + \pi^*(q)$.
\textbf{Step 5:} Finally, $\Upsilon_i$ is calculated by:
\begin{equation}\label{eq_werdh}
\Upsilon_i = \alpha \cdot \mathrm{cos}(z_i,\bar{Z}_i - \bar{Z'}_i + \bar{z}),
\end{equation}
where $\alpha$ is a scaling hyperparameter, and $\mathrm{cos}(\cdot, \cdot)$ denotes the cosine similarity between two vectors.

\textbf{Design of KL Divergence}. The design of $D_{\rm KL}(\pi_{\theta} \parallel \pi'_{\rm ref})$ is directly inspired by Theorem~\ref{pro_dsf}. According to Equation~(\ref{eq_asgfh}), the expected risk of $\pi^*(x)$ is higher than that of $\Psi \cdot \pi^*(x) + \pi^*(q)$. This suggests that the output generated by $\Psi \cdot \pi^*(x) + \pi^*(q)$ may have better quality than the one produced by $\pi^*(x)$. Since $\pi$ represents the probability distribution over output tokens, while $\pi^*$ corresponds to the actual generated content, a natural way to improve the performance of $\pi$ is to encourage its output distribution to align with that of $\Psi \cdot \pi(x) + \pi(q)$. This motivates the use of KL divergence as a regularization term. Specifically, for $D_{\rm KL}(\pi_{\theta} \parallel \pi'_{\rm ref})$, the definition follows a procedure similar to Equation (\ref{eq_gfdgrpo}):
\begin{equation}\label{eq_gfqrpo}
\sum\limits_{i=0}^{n-1} [\frac{\pi'_{\rm ref}(x_i)}{\pi_{\theta}(y_{i,j}|x_i,y_{i,<j})} -\log \frac{\pi'_{\rm ref}(x_i)}{\pi_{\theta}(y_{i,j}|x_i,y_{i,<j})}-1],
\end{equation} 
where $\pi'_{\rm ref}(x_i) = \Psi \cdot \pi(y_{i,j}|x_i,y_{i,<j}) + \pi(y_{i,j}|q,y_{i,<j})$. Based on the analysis in the previous paragraph, we derive the following approximations:
(1) $\Phi \cdot \pi(y_{i,j} \mid x_i,y_{i,<j})$ can be approximated by
$
\sum_{l=0}^{n-1}\pi(y_{i,j} \mid x_i,y_{n,l},y_{i,<j})
$,
and (2) the analytical expression of $\Psi \cdot \pi(y_{i,j} \mid x_i, y_{i,<j}) + \pi(y_{i,j} \mid q, y_{i,<j})$ can be approximated by the following equation:
\begin{equation}\label{eq_gfqrsgdfhpo}
\begin{array}{c}
\pi(y_{i,j} \mid x_i,y_{i,<j}) - \sum\limits_{l=0}^{n-1} 
\pi(y_{i,j} \mid x_i,y_{n,l},y_{i,<j}) \\
+ \pi(y_{i,j} \mid q, y_{i,<j}).
\end{array}
\end{equation}
Note that, similar to GRPO, the computation in Equation (\ref{eq_gfqrpo}) is also carried out in a token-wise manner. Finally, the training process is also similar to GRPO. In the appendix, the overall procedure of the GCPO training is illustrated through the pseudocode.

\begin{table*}[t]
  \centering
  \small
    \begin{tabular}{l|c|c|c|c|c|c}
      \toprule
      Base model + Method & AIME 2024 & AIME 2025 & AMC 2023 & MATH500 & MinervaMATH & \textbf{Avg.} \\
      \midrule
      \!\textbf{DeepScaleR‑1.5B‑Preview} & 42.8 & 36.7 & 83.0 & 85.2 & 24.6 & 54.5 \\
      \!~~+GRPO \cite{shao2024deepseekmath} & 44.5 (+1.7) & 39.3 (+2.6) & 81.5 (-1.5) & 84.9 (-0.3) & 24.7 (+0.1) & 55.0 (+0.5) \\
      \!~~+ReST-MCTS \cite{zhang2024rest}& 45.5 (+2.7) & 39.5 (+2.8) & 83.4 (+0.4) & 84.8 (-0.4) & 23.9 (-0.7) & 55.4 (+0.9) \\
      \!~~+GVPO \cite{zhang2025gvpo} & 46.1 (+3.3) & 39.7 (+3.0) & 83.6 (+0.6) & 85.7 (+0.5) & 25.3 (+0.7) & 56.1 (+1.6) \\
      \!~~+Dr.GRPO \cite{liu2025understanding} & 45.8 (+3.0) & 39.6 (+2.9) & 82.1 (-0.9) & 85.3 (+0.1) & 25.1 (+0.5) & 55.6 (+1.1) \\
       \!~~+GCPO (Ours) & \textbf{46.7 (+3.9)} & \textbf{40.3 (+3.6)} & \textbf{84.1 (+1.1)} & \textbf{86.3 (+1.1)} & \textbf{25.9 (+1.4)} & \textbf{56.8 (+2.3)} \\
      \midrule
      \!\textbf{DeepSeek-R1-Distill-Qwen-1.5B} & 28.7 & 26.0 & 69.9 & 80.1 & 19.8 & 44.9 \\
      \!~~+GRPO \cite{shao2024deepseekmath} & 29.8 (+1.1) & 27.3 (+1.3) & 70.5 (+0.6) & 80.3 (+0.2) & 22.1 (+2.3) & 46.0 (+1.1) \\
      \!~~+ReST-MCTS \cite{zhang2024rest} & 30.5 (+1.8) & 28.6 (+2.6) & 71.1 (+1.2) & 80.4 (+0.3) & 20.3 (+0.5) & 46.4 (+1.5) \\
      \!~~+GVPO \cite{zhang2025gvpo} & 30.6 (+1.9) & 28.2 (+2.2) & 71.5 (+1.6) & 80.5 (+0.4) & 23.1 (+3.3) & 46.7 (+1.8) \\
      \!~~+Dr.GRPO \cite{liu2025understanding} & 30.4 (+1.7) & 28.4 (+2.4) & 71.3 (+1.4) & 80.8 (+0.7) & 22.9 (+3.1) & 46.9 (+2.0) \\
      \!~~+GCPO (Ours) & \textbf{31.0 (+2.3)} & \textbf{29.0 (+3.0)} & \textbf{71.8 (+1.9)} & \textbf{81.6 (+1.5)} & \textbf{23.4 (+3.6)} & \textbf{47.4 (+2.5)} \\
         \midrule
    \!\textbf{DeepSeek-R1-Distill-Qwen-7B} & 55.5 & 50.2 & 85.1 & 87.4 & 42.1 & 64.1 \\
    \!~~+GRPO \cite{shao2024deepseekmath} & 56.9 (+1.4) & 51.7 (+1.5) & 85.5 (+0.4) & 87.7 (+0.3) & 43.5 (+1.4) & 65.1 (+1.0) \\
    \!~~+ReST-MCTS \cite{zhang2024rest}  & 57.1 (+1.6) & 52.4 (+2.2) & 85.7 (+0.6) & 87.9 (+0.5) & 42.8 (+0.7) & 65.2 (+1.1) \\
    \!~~+GVPO \cite{zhang2025gvpo}       & 57.5 (+2.0) & 52.1 (+1.9) & 86.3 (+1.2) & 88.5 (+1.1) & 44.2 (+2.1) & 65.7 (+1.6) \\
    \!~~+Dr.GRPO \cite{liu2025understanding} & 57.4 (+1.9) & 52.3 (+2.1) & 86.4 (+1.3) & 88.2 (+0.8) & 44.0 (+1.9) & 65.7 (+1.6) \\
    \!~~+GCPO (Ours) & \textbf{58.3 (+2.8)} & \textbf{53.0 (+2.8)} & \textbf{87.3 (+2.2)} & \textbf{89.1 (+1.7)} & \textbf{45.0 (+2.9)} & \textbf{66.5 (+2.4)} \\
      \bottomrule
    \end{tabular}%
  \caption{Pass@1 performance on various math reasoning benchmarks. We compare base models trained with different fine-tuning approaches. The best results are highlighted in \textbf{bold}.}
  \label{tab:ex_1}
\end{table*}



\begin{figure*}[t]
    \centering
    \begin{minipage}[t]{0.235\linewidth}
        \centering
        \includegraphics[width=\linewidth]{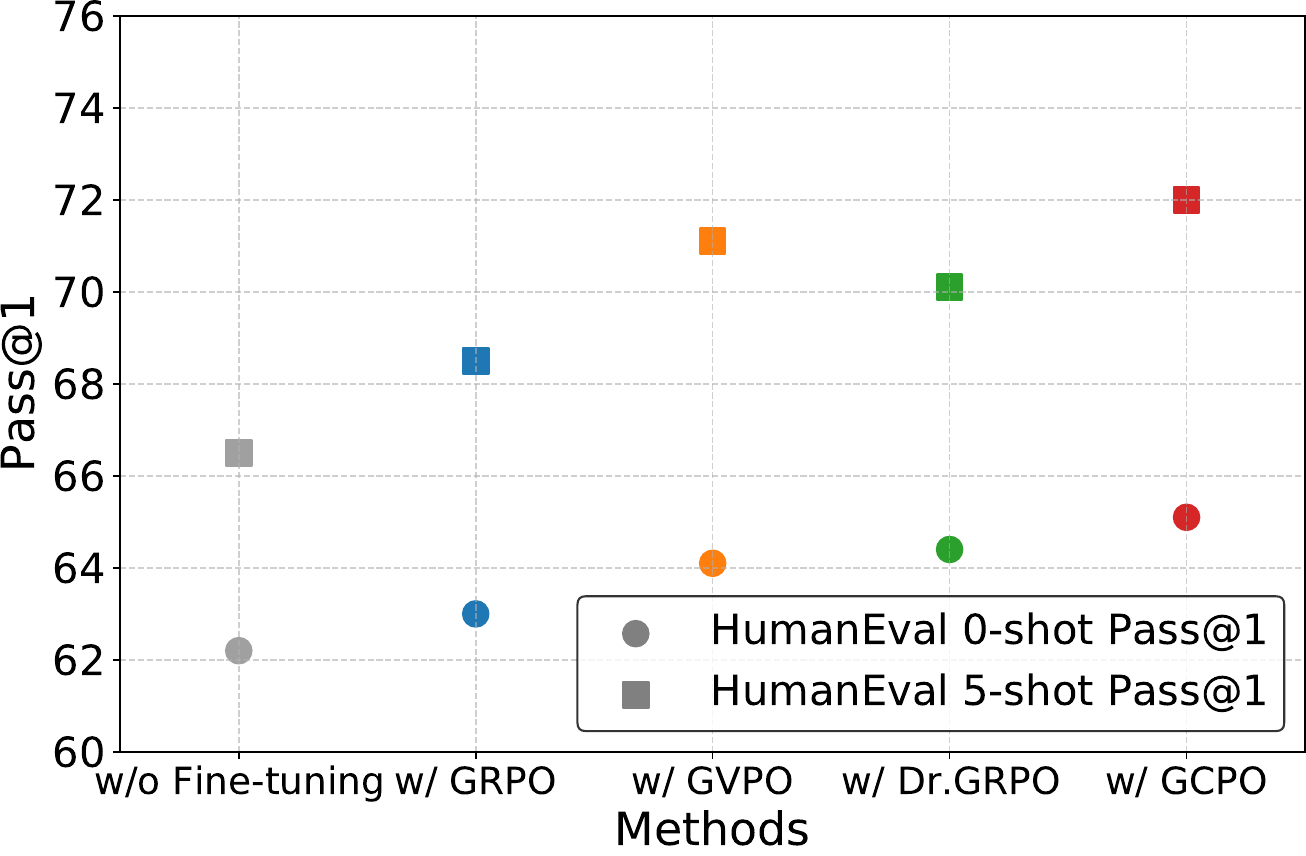}
        \caption{Performance analysis on code reasoning tasks. We record the 1-shot and 5-shot results on HumanEval.}
        \label{fig:performance_code}
    \end{minipage}
    \hfill
    \begin{minipage}[t]{0.755\linewidth}
        \centering
    \subfigure[Ablation Study of $\alpha$]{
        \includegraphics[width=0.32\linewidth]{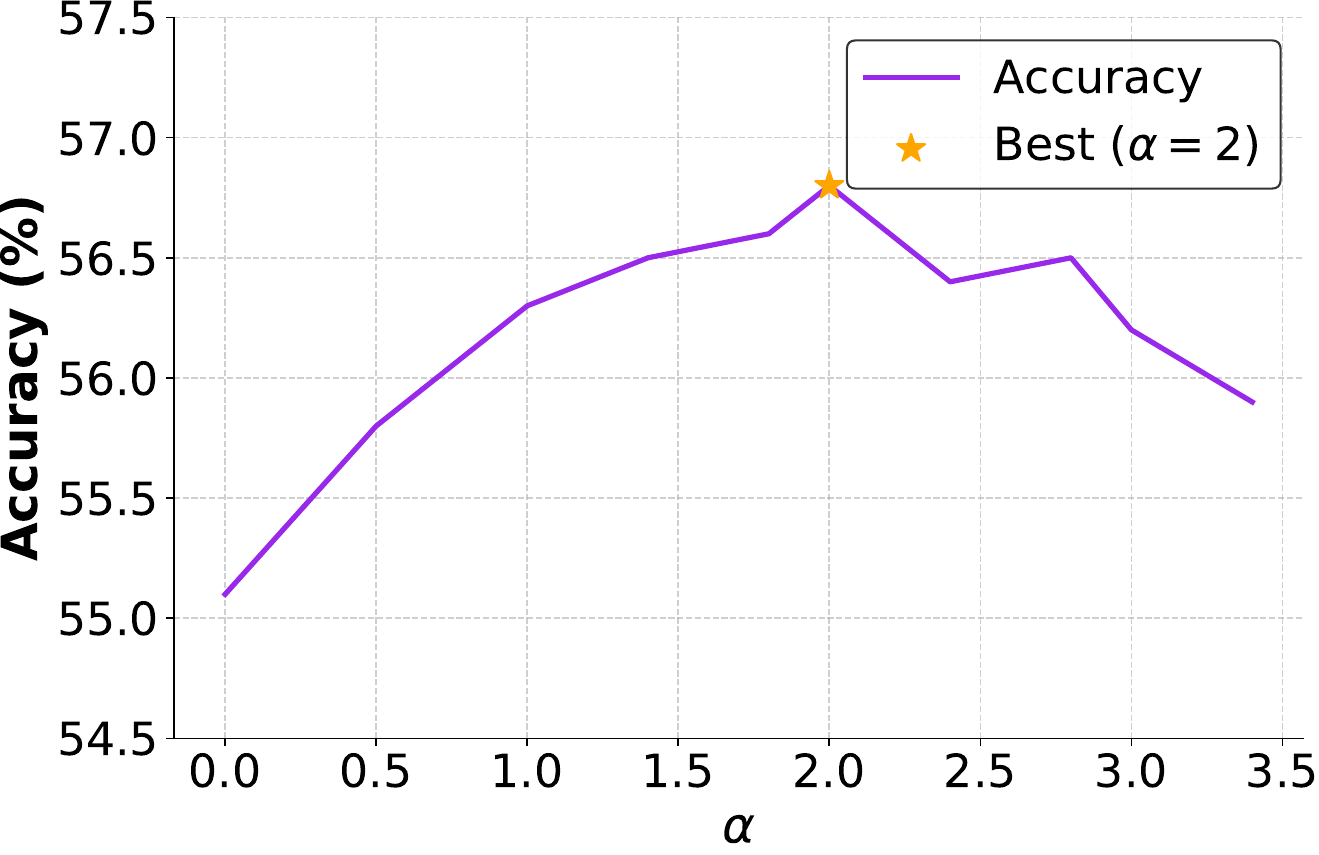}}
        \hfill
    \subfigure[Ablation Study of $\kappa$]{
        \includegraphics[width=0.32\linewidth]{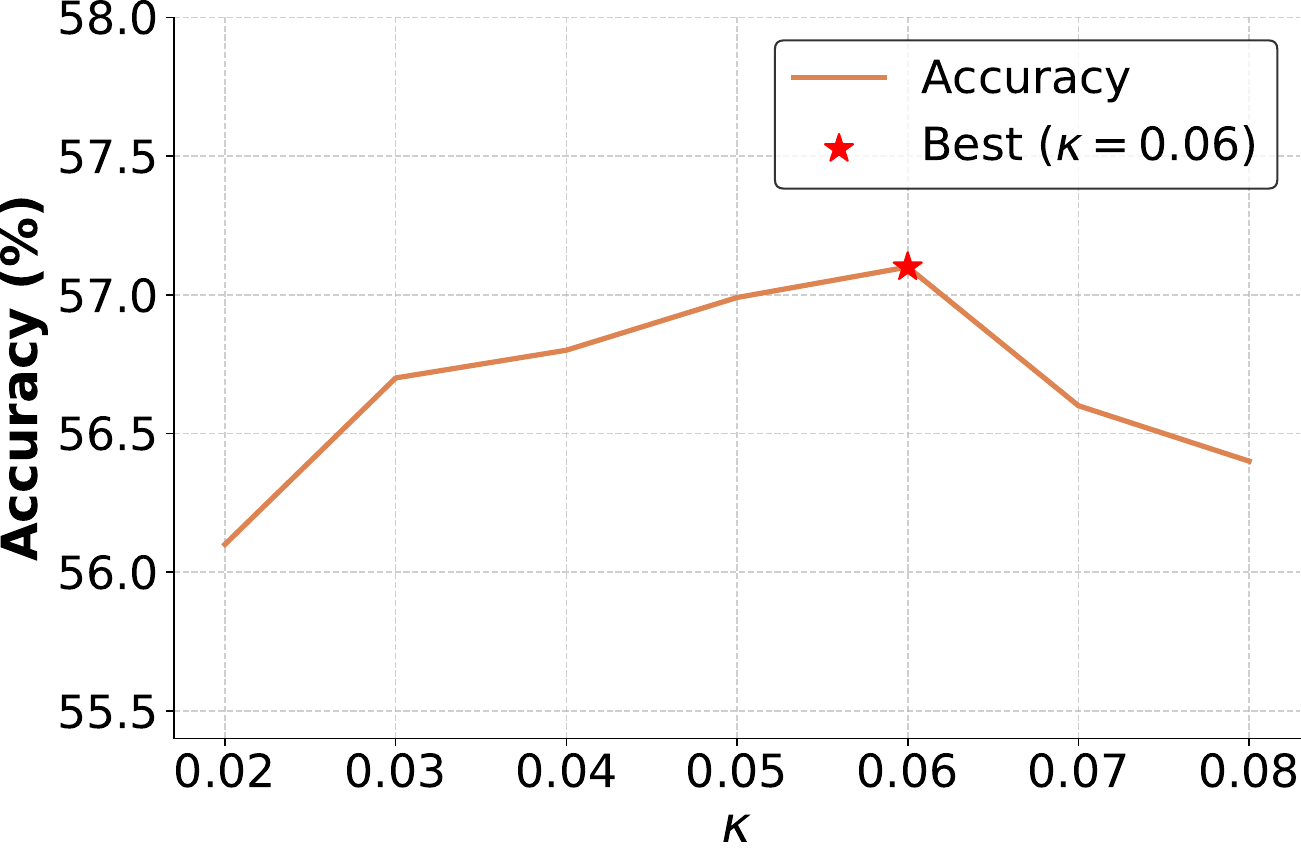}}
    \subfigure[Evaluation of similarity metric]{
        \includegraphics[width=0.32\linewidth]{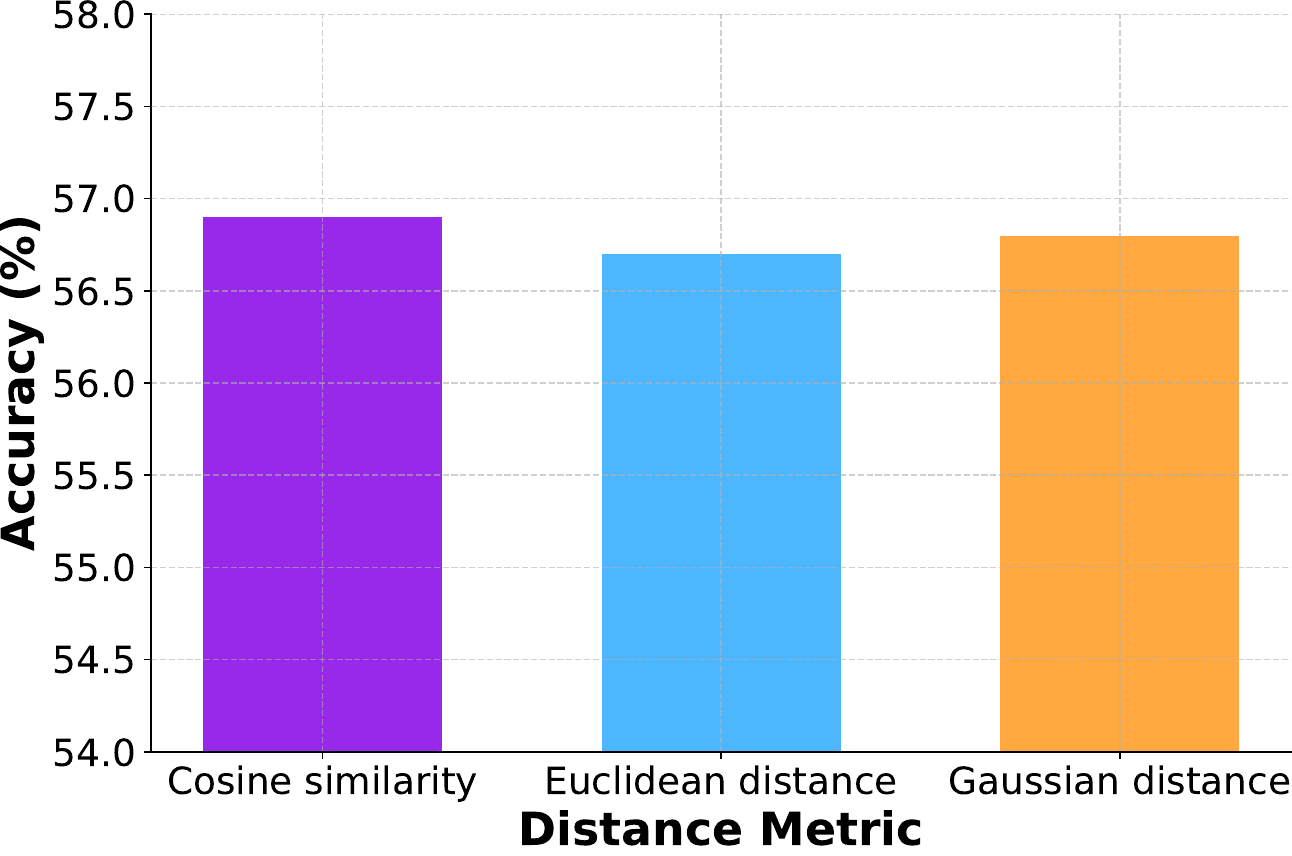}}
    \caption{Ablation Study. (a) and (b) shows the results for parameter sensitivity, i.e., hyperparameter $\alpha$ and $\kappa$. (c) shows the evaluation of similarity metric in $\Upsilon_i$.}
    \label{fig:ex_2}
    \end{minipage}
\end{figure*}


\begin{table*}[t]
\centering
\small
\begin{tabular}{lccccc|c}
\toprule
\textbf{Configurations of GCPO} & AIME 2024 & AIME 2025 & AMC 2023 & MATH500 & MinervaMATH & \textbf{Avg.} \\
\midrule
\quad GCPO (Full) & 46.7 & 40.3 & 84.1 & 86.3 & 25.9 & 56.8 \\
\quad w/o Adv. Weighting & 45.0 & 39.0 & 83.0 & 85.4 & 24.6 & 55.4 \\
\quad w/o KL Term & 45.3 & 39.4 & 83.3 & 85.3 & 24.9 & 55.6 \\
\bottomrule
\end{tabular}
\caption{Ablation study of the components within GCPO. We report Pass@1 on five benchmarks. Removing either the advantage weighting or KL divergence degrades overall performance, indicating both are essential.}
\label{tab:ablation_com}
\end{table*}


\section{Experiments}
In this section, we conduct comprehensive experiments and ablation studies on multiple reasoning benchmarks to evaluate the effectiveness of our proposed method. 

\subsection{Experimental Settings}
We conduct evaluations of our method across several reasoning benchmarks, including AIME24-25, AMC, MATH500 \cite{hendrycks2021measuring}, MinervaMATH \cite{lewkowycz2022solving}, and HumanEval \cite{chen2021evaluating}. Our experiments use DeepScaleR-1.5B-Preview and DeepSeek-R1-Distill-Qwen-1.5B, and DeepSeekR1-Distill-Qwen-7B, and Qwen2-7B-Instruct as base models. We compare our method against classic and  SOTA reinforcement learning methods, including GRPO \cite{shao2024deepseekmath}, GVPO \cite{zhang2025gvpo}, ReST-MCTS \cite{zhang2024rest}, and Dr.GRPO \cite{liu2025understanding}.
DeepScaleR-1.5B-Preview, having been previously fine-tuned on 40k math QA pairs, is further fine-tuned on 919 AIME problems from 1989 to 2023. DeepSeek-R1-Distill-Qwen-1.5B is fine-tuned on a random subset of 4,000 QA pairs from NuminaMath \cite{li2024numinamath}. Following \cite{wang2025learning}, all training and evaluation stages are constrained to a token budget of 16,384. We adopt a learning rate of $1\mathrm{e}{-6}$, weight decay of 0.01, and batch size of 256.
All experiments are conducted on A100 GPU clusters.

\subsection{Performance Analysis}

\subsubsection{Performance on Mathematical Reasoning Tasks.}
We evaluate our method against all baseline approaches across a comprehensive set of benchmarks, including AIME 2024, AIME 2025, AMC 2023, MATH500, and MinervaMATH. We compare three widely used base models, including DeepScaleR-1.5B-Preview, DeepSeek-R1-Distill-Qwen-1.5B, and DeepSeek-R1-Distill-Qwen-7B. We report the pass@1 accuracy following \cite{wang2025learning}. Table \ref{tab:ex_1} shows the pass@1 performance. From the results, we can observe that across all settings, GCPO consistently achieves the best average performance, outperforming both the base models and all competitive baselines. Specifically, for DeepScaleR-1.5B-Preview, GCPO delivers an average improvement of 2.3\% over the base model. Similar trends are observed for DeepSeek-R1-Distill-Qwen-1.5B and DeepSeek-R1-Distill-Qwen-7B, where GCPO yields average improvements of 2.5\% and 2.2\%, respectively.
Compared to SOTA RL baselines such as GRPO and GVPO, GCPO offers stronger gains on more challenging datasets (e.g., over 1\% on AIME and MinervaMATH), highlighting its effectiveness in capturing complex reasoning patterns through causally informed optimization. Notably, the performance margins between GCPO and the strongest baseline are largest on the hardest benchmarks, indicating that the benefits of causal group structure modeling become more pronounced as task complexity increases.
Moreover, the consistent improvements demonstrate that GCPO is robust to model scale and architecture, providing a generalizable fine-tuning strategy for math reasoning tasks.

\subsubsection{Performance on Code Reasoning Tasks.}
Further, for code reasoning tasks, we run different fine-tuning pipelines on Qwen2-7B-Instruct and evaluate them using the standard HumanEval protocol. The results are summarized in Figure~\ref{fig:performance_code}, which reports both 0-shot and 5-shot Pass@1 accuracies for each method. From the results, we can observe that across both evaluation settings, GCPO achieves the strongest performance among all compared methods. In particular, GCPO attains a 0-shot Pass@1 of 65.1\% and a 5-shot Pass@1 of 72.0\%, outperforming the foundation model by 2.9\% and 5.5\%, respectively. Relative to other policy optimization methods, such as GRPO and GVPO, GCPO consistently delivers higher accuracy. Notably, the gap between GCPO and previous methods becomes even more pronounced in the multi-shot evaluation, highlighting the advantage of incorporating causal projection and structure-aware regularization in leveraging contextual information and enabling compositional reasoning. These results further demonstrate the effectiveness of the proposed GCPO.

\subsubsection{Visualization Analysis}
Given the substantial computational cost of training LLMs, maintaining stable training dynamics is crucial. To assess this, we use the gradient norm (as a proxy for policy variance) to measure training stability. We record the gradient norm during training for both the baseline methods and the proposed GCPO. The results, presented in the Appendix (Figure 1), demonstrate that GCPO achieves the highest stability, with the gradient norm remaining consistently steady throughout training.

\subsection{Ablation Study}
We conduct a series of ablation studies to evaluate the contribution of each component within our method, the best parameterization and implementation choices, etc.

\subsubsection{The effect of different components.}
To assess the effect of each component in GCPO, we conduct ablation studies. Specifically, we consider two alternative configurations: (i) removing the advantage weighting term (i.e., reusing $A_i$ for all $i$); and (ii) removing the additional KL divergence term (i.e., setting $\kappa = 0$). Notably, the overall contribution of our reward formulation has already been substantiated in Table~\ref{tab:ex_1}. Here, we focus on isolating the impact of these individual mechanisms.
The ablation results are shown in Figure \ref{tab:ablation_com}. We can observe that both terms are critical for LLM reasoning. These findings underscore the advantages of our design and the effectiveness of GCPO.

\subsubsection{Parameter sensitivity.}
We select the hyperparameters of GCPO based on a systematic evaluation of reasoning performance. Specifically, we conduct a grid search over the hyperparameter $\alpha$ and the KL regularization coefficient $\kappa$ to identify the optimal configuration for our method. For $\alpha$, which controls the influence of the causal projection factor, we explore a range of values: $[0, 0.5, 1, 1.4, 1.8, 2, 2.4, 2.8, 3, 3.4]$. 
For $\kappa$, which balances the KL regularization strength, we first perform a coarse grid search over $[0.02, 0.04, 0.06, 0.08]$ with a step size of $0.02$, and subsequently conduct a finer search within the promising interval using a step size of $0.01$.
For each configuration of $(\alpha, \beta)$, we record the Pass@1 performance. As shown in Figure~\ref{fig:ex_2}(a)-(b), model accuracy initially increases with larger values of $\alpha$ and $\kappa$, but plateaus or slightly degrades when these values become too large; the best results are consistently achieved with $\alpha=2$ and $\beta=0.06$. These values are thus adopted as our default hyperparameter settings.

\subsubsection{Evaluation of metric for $\Upsilon_i$} 
According to Eq.~\ref{eq_werdh}, we compute $\Upsilon_i$ by calculating the cosine similarity $cos(\cdot)$. To evaluate the impact of this metric on performance, we conduct an ablation study comparing different similarity measures, including cosine similarity, Euclidean distance, and Gaussian distance. The results are shown in Figure~\ref{fig:ex_2}(c). With the introduction of the hyperparameter $\alpha$, the performance differences among various similarity measures are negligible. We ultimately select cosine similarity as the default metric, primarily because it allows for more flexible and convenient tuning of $\alpha$ (See the Appendix for details).

\section{Conclusion}

In this paper, we present GCPO, a novel post-training method that integrates causal structure into policy optimization for large language models. Building on the limitations of GRPO, GCPO addresses the overlooked interdependencies among groupwise candidate responses by modeling them through an SCM. Our analysis reveals that conditioning on a final integrated response induces a collider structure, which in turn exposes latent dependencies among originally independent candidates. Guided by this insight, GCPO introduces two key components: (1) a causally-adjusted reward mechanism that projects individual responses onto a structurally coherent subspace, and (2) a KL-divergence regularization term that aligns the policy with a causally-informed reference distribution. Extensive experiments across multiple reasoning benchmarks demonstrate that GCPO substantially outperforms existing baselines, confirming the benefits of incorporating causal reasoning into groupwise optimization. Our findings underscore the importance of structural awareness in reinforcement learning for LLM post-training and suggest promising directions for future work on causality-aware RLHF.

\bibliography{aaai2026}

\begin{thebibliography}{36}
\providecommand{\natexlab}[1]{#1}

\bibitem[{Bai et~al.(2024)Bai, Zhou, Cemri, Pan, Suhr, Levine, and Kumar}]{bai2024digirltraininginthewilddevicecontrol}
Bai, H.; Zhou, Y.; Cemri, M.; Pan, J.; Suhr, A.; Levine, S.; and Kumar, A. 2024.
\newblock DigiRL: Training In-The-Wild Device-Control Agents with Autonomous Reinforcement Learning.
\newblock arXiv:2406.11896.

\bibitem[{Bai et~al.(2022)Bai, Kadavath, Kundu, Askell, Kernion, Jones, Chen, Goldie, Mirhoseini, McKinnon et~al.}]{bai2022training}
Bai, Y.; Kadavath, S.; Kundu, S.; Askell, A.; Kernion, J.; Jones, A.; Chen, A.; Goldie, A.; Mirhoseini, A.; McKinnon, C.; et~al. 2022.
\newblock Training a Helpful and Harmless Assistant with Reinforcement Learning from Human Feedback.
\newblock \emph{arXiv preprint arXiv:2204.05862}.

\bibitem[{Ballon, Algaba, and Ginis(2025)}]{ballon2025relationship}
Ballon, M.; Algaba, A.; and Ginis, V. 2025.
\newblock The Relationship Between Reasoning and Performance in Large Language Models--o3 (mini) Thinks Harder, Not Longer.
\newblock \emph{arXiv preprint arXiv:2502.15631}.

\bibitem[{Bengio et~al.(2003)Bengio, Ducharme, Vincent, and Jauvin}]{bengio2003neural}
Bengio, Y.; Ducharme, R.; Vincent, P.; and Jauvin, C. 2003.
\newblock A neural probabilistic language model.
\newblock \emph{Journal of machine learning research}, 3(Feb): 1137--1155.

\bibitem[{Bottou, Curtis, and Nocedal(2018)}]{bottou2018optimizationmethodslargescalemachine}
Bottou, L.; Curtis, F.~E.; and Nocedal, J. 2018.
\newblock Optimization Methods for Large-Scale Machine Learning.
\newblock arXiv:1606.04838.

\bibitem[{Chen et~al.(2021)Chen, Tworek, Jun, Yuan, Pinto, Kaplan, Edwards, Burda, Joseph, Brockman et~al.}]{chen2021evaluating}
Chen, M.; Tworek, J.; Jun, H.; Yuan, Q.; Pinto, H. P. D.~O.; Kaplan, J.; Edwards, H.; Burda, Y.; Joseph, N.; Brockman, G.; et~al. 2021.
\newblock Evaluating large language models trained on code.
\newblock \emph{arXiv preprint arXiv:2107.03374}.

\bibitem[{Cheng et~al.(2025)Cheng, Chen, Fu, and Zhou}]{cheng2025optimizing}
Cheng, Z.; Chen, D.; Fu, M.; and Zhou, T. 2025.
\newblock Optimizing Length Compression in Large Reasoning Models.
\newblock \emph{arXiv preprint arXiv:2506.14755}.

\bibitem[{Chung et~al.(2022)Chung, Hou, Longpre, Zoph, Tay, Fedus, Li, Wang, Hu, Roberts, Mehta, Wei, Chandu, Gritsenko, Piantanida, Chowdhery, Clark, Schick, Dwivedi-Yu, Yu, Shi, Li, Ippolito, Zhou, Ainslie, Firat, Lu, Dean, Le, and Chi}]{chung2022scaling}
Chung, H.~W.; Hou, L.; Longpre, S.; Zoph, B.; Tay, Y.; Fedus, W.; Li, E.; Wang, X.; Hu, X.; Roberts, A.; Mehta, H.; Wei, J.; Chandu, K.~R.; Gritsenko, A.; Piantanida, P.; Chowdhery, A.; Clark, J.~H.; Schick, T.; Dwivedi-Yu, J.; Yu, J.; Shi, K.; Li, X.; Ippolito, D.; Zhou, D.; Ainslie, J.; Firat, O.; Lu, Y.; Dean, J.; Le, Q.~V.; and Chi, E.~H. 2022.
\newblock Scaling Instruction-Finetuned Language Models.
\newblock \emph{arXiv preprint arXiv:2210.11416}.

\bibitem[{Devlin et~al.(2019)Devlin, Chang, Lee, and Toutanova}]{devlin2019bert}
Devlin, J.; Chang, M.-W.; Lee, K.; and Toutanova, K. 2019.
\newblock BERT: Pre-training of Deep Bidirectional Transformers for Language Understanding.
\newblock In \emph{Proceedings of the 2019 Conference of the North American Chapter of the Association for Computational Linguistics: Human Language Technologies, Volume 1 (Long and Short Papers)}, 4171--4186.

\bibitem[{Goodfellow, Bengio, and Courville(2016)}]{goodfellow2016deep}
Goodfellow, I.; Bengio, Y.; and Courville, A. 2016.
\newblock \emph{Deep learning}, volume~1.

\bibitem[{Guo et~al.(2025)Guo, Yang, Zhang, Song, Zhang, Xu, Zhu, Ma, Wang, Bi et~al.}]{guo2025deepseek}
Guo, D.; Yang, D.; Zhang, H.; Song, J.; Zhang, R.; Xu, R.; Zhu, Q.; Ma, S.; Wang, P.; Bi, X.; et~al. 2025.
\newblock Deepseek-r1: Incentivizing reasoning capability in llms via reinforcement learning.
\newblock \emph{arXiv preprint arXiv:2501.12948}.

\bibitem[{Hendrycks et~al.(2021)Hendrycks, Burns, Kadavath, Arora, Basart, Tang, Song, and Steinhardt}]{hendrycks2021measuring}
Hendrycks, D.; Burns, C.; Kadavath, S.; Arora, A.; Basart, S.; Tang, E.; Song, D.; and Steinhardt, J. 2021.
\newblock Measuring mathematical problem solving with the math dataset.
\newblock \emph{arXiv preprint arXiv:2103.03874}.

\bibitem[{Jaech et~al.(2024)Jaech, Kalai, Lerer, Richardson, El-Kishky, Low, Helyar, Madry, Beutel, Carney et~al.}]{jaech2024openai}
Jaech, A.; Kalai, A.; Lerer, A.; Richardson, A.; El-Kishky, A.; Low, A.; Helyar, A.; Madry, A.; Beutel, A.; Carney, A.; et~al. 2024.
\newblock Openai o1 system card.
\newblock \emph{arXiv preprint arXiv:2412.16720}.

\bibitem[{Lai et~al.(2025)Lai, Liu, Gao, Cheng, Qi, Xu, Yao, Zhang, Du, Hou et~al.}]{lai2025survey}
Lai, H.; Liu, X.; Gao, J.; Cheng, J.; Qi, Z.; Xu, Y.; Yao, S.; Zhang, D.; Du, J.; Hou, Z.; et~al. 2025.
\newblock A Survey of Post-Training Scaling in Large Language Models.
\newblock In \emph{Proceedings of the 63rd Annual Meeting of the Association for Computational Linguistics (Volume 1: Long Papers)}, 2771--2791.

\bibitem[{Lewkowycz et~al.(2022)Lewkowycz, Andreassen, Dohan, Dyer, Michalewski, Ramasesh, Slone, Anil, Schlag, Gutman-Solo et~al.}]{lewkowycz2022solving}
Lewkowycz, A.; Andreassen, A.; Dohan, D.; Dyer, E.; Michalewski, H.; Ramasesh, V.; Slone, A.; Anil, C.; Schlag, I.; Gutman-Solo, T.; et~al. 2022.
\newblock Solving quantitative reasoning problems with language models.
\newblock \emph{Advances in Neural Information Processing Systems}, 35: 3843--3857.

\bibitem[{Li et~al.(2024)Li, Beeching, Tunstall, Lipkin, Soletskyi, Huang, Rasul, Yu, Jiang, Shen et~al.}]{li2024numinamath}
Li, J.; Beeching, E.; Tunstall, L.; Lipkin, B.; Soletskyi, R.; Huang, S.; Rasul, K.; Yu, L.; Jiang, A.~Q.; Shen, Z.; et~al. 2024.
\newblock Numinamath: The largest public dataset in ai4maths with 860k pairs of competition math problems and solutions.
\newblock \emph{Hugging Face repository}, 13: 9.

\bibitem[{Liu et~al.(2025)Liu, Chen, Li, Qi, Pang, Du, Lee, and Lin}]{liu2025understanding}
Liu, Z.; Chen, C.; Li, W.; Qi, P.; Pang, T.; Du, C.; Lee, W.~S.; and Lin, M. 2025.
\newblock Understanding r1-zero-like training: A critical perspective.
\newblock \emph{arXiv preprint arXiv:2503.20783}.

\bibitem[{Minaee et~al.(2024{\natexlab{a}})Minaee, Mikolov, Nikzad, Chenaghlu, Socher, Amatriain, and Gao}]{minaee2024survey}
Minaee, S.; Mikolov, T.; Nikzad, N.; Chenaghlu, M.; Socher, R.; Amatriain, X.; and Gao, J. 2024{\natexlab{a}}.
\newblock Large Language Models: A Survey.
\newblock \emph{arXiv preprint arXiv:2402.06196}.

\bibitem[{Minaee et~al.(2024{\natexlab{b}})Minaee, Mikolov, Nikzad, Chenaghlu, Socher, Amatriain, and Gao}]{minaee2024large}
Minaee, S.; Mikolov, T.; Nikzad, N.; Chenaghlu, M.; Socher, R.; Amatriain, X.; and Gao, J. 2024{\natexlab{b}}.
\newblock Large language models: a survey (2024).
\newblock \emph{URL https://arxiv. org/abs/2402.06196}, 7(8): 9.

\bibitem[{Muennighoff et~al.(2025)Muennighoff, Yang, Shi, Li, Fei-Fei, Hajishirzi, Zettlemoyer, Liang, Cand{\`e}s, and Hashimoto}]{muennighoff2025s1}
Muennighoff, N.; Yang, Z.; Shi, W.; Li, X.~L.; Fei-Fei, L.; Hajishirzi, H.; Zettlemoyer, L.; Liang, P.; Cand{\`e}s, E.; and Hashimoto, T. 2025.
\newblock s1: Simple test-time scaling.
\newblock \emph{arXiv preprint arXiv:2501.19393}.

\bibitem[{Ouyang et~al.(2022)Ouyang, Wu, Jiang, Almeida, Wainwright, Mishkin, Zhang, Agarwal, Slama, Ray et~al.}]{ouyang2022training}
Ouyang, L.; Wu, J.; Jiang, X.; Almeida, D.; Wainwright, C.; Mishkin, P.; Zhang, C.; Agarwal, S.; Slama, K.; Ray, A.; et~al. 2022.
\newblock Training language models to follow instructions with human feedback.
\newblock \emph{arXiv preprint arXiv:2203.02155}.

\bibitem[{Pearl(2009)}]{pearl2009causality}
Pearl, J. 2009.
\newblock \emph{Causality}.
\newblock Cambridge university press.

\bibitem[{Pearl, Glymour, and Jewell(2016)}]{pearl2016causal}
Pearl, J.; Glymour, M.; and Jewell, N.~P. 2016.
\newblock \emph{Causal inference in statistics: A primer}.
\newblock John Wiley \& Sons.

\bibitem[{Raffel et~al.(2020)Raffel, Shazeer, Roberts, Lee, Narang, Matena, Zhou, Li, and Liu}]{raffel2020exploring}
Raffel, C.; Shazeer, N.; Roberts, A.; Lee, K.; Narang, S.; Matena, M.; Zhou, Y.; Li, W.; and Liu, P.~J. 2020.
\newblock Exploring the Limits of Transfer Learning with a Unified Text-to-Text Transformer.
\newblock \emph{Journal of Machine Learning Research}, 21(140): 1--67.

\bibitem[{Sadik and Govind(2025)}]{sadik2025benchmarkingllmcodesmells}
Sadik, A.~R.; and Govind, S. 2025.
\newblock Benchmarking LLM for Code Smells Detection: OpenAI GPT-4.0 vs DeepSeek-V3.
\newblock arXiv:2504.16027.

\bibitem[{Sanh et~al.(2022)Sanh, Webson, Raffel, Bach, Sutawika, Alyafeai, Chaffin, Srinivasan, Yong, Kim, Crowell, Kudugunta, Sharma, Ong, Sharma, Lo, Bari, Xu, Thakker, Dey, Desai, Sangwan, Geng, Arora, Ram, Wang, Chandu, Kashyap, Tan, Gotmare, Swabha, Phang, Chan, Urbanek, Gururangan, d.~S.~Clemente, McMahan, Albanie, Welbl, Liu, Malmi, Jean, Kuo, Jiang, Xu, Conneau, McCoy, Taylor, Smith, Zettlemoyer, Ruder, Yogatama, Cho, and Rush}]{sanh2022multitask}
Sanh, V.; Webson, A.; Raffel, C.; Bach, S.~H.; Sutawika, L.; Alyafeai, Z.; Chaffin, A.; Srinivasan, A.; Yong, Z.~X.; Kim, T.; Crowell, E.~S.; Kudugunta, S.; Sharma, A.; Ong, R.; Sharma, S.; Lo, A.; Bari, M.~S.; Xu, C.; Thakker, U.; Dey, M.; Desai, S.; Sangwan, R.; Geng, X.; Arora, D.; Ram, D.; Wang, H.; Chandu, K.; Kashyap, A.; Tan, S.; Gotmare, A.~D.; Swabha, S.; Phang, J.; Chan, H.~P.; Urbanek, J.~H.; Gururangan, S.; d.~S.~Clemente, M.~V.; McMahan, B.; Albanie, S.; Welbl, J.; Liu, Q.; Malmi, E.; Jean, S.; Kuo, J.~T.; Jiang, M. T.-J.; Xu, Y.; Conneau, A.; McCoy, R.~T.; Taylor, S.; Smith, N.~A.; Zettlemoyer, L.; Ruder, S.; Yogatama, D.; Cho, K.; and Rush, A.~M. 2022.
\newblock Multitask Prompted Training Enables Zero-Shot Task Generalization.
\newblock In \emph{International Conference on Learning Representations (ICLR)}.

\bibitem[{Schulman et~al.(2017)Schulman, Wolski, Dhariwal, Radford, and Klimov}]{schulman2017proximal}
Schulman, J.; Wolski, F.; Dhariwal, P.; Radford, A.; and Klimov, O. 2017.
\newblock Proximal policy optimization algorithms.
\newblock \emph{arXiv preprint arXiv:1707.06347}.

\bibitem[{Shao et~al.(2024)Shao, Wang, Zhu, Xu, Song, Bi, Zhang, Zhang, Li, Wu et~al.}]{shao2024deepseekmath}
Shao, Z.; Wang, P.; Zhu, Q.; Xu, R.; Song, J.; Bi, X.; Zhang, H.; Zhang, M.; Li, Y.; Wu, Y.; et~al. 2024.
\newblock Deepseekmath: Pushing the limits of mathematical reasoning in open language models.
\newblock \emph{arXiv preprint arXiv:2402.03300}.

\bibitem[{Tie et~al.(2025)Tie, Zhao, Song, Wei, Zhou, Dai, Yin, Yang, Yan, Su et~al.}]{tie2025survey}
Tie, G.; Zhao, Z.; Song, D.; Wei, F.; Zhou, R.; Dai, Y.; Yin, W.; Yang, Z.; Yan, J.; Su, Y.; et~al. 2025.
\newblock A survey on post-training of large language models.
\newblock \emph{arXiv e-prints}, arXiv--2503.

\bibitem[{Touvron et~al.(2023)Touvron, Lavril, Izacard, Martinet, Lachaux, Lacroix, Rozi{\`e}re, Goyal, Hambro, Azhar et~al.}]{touvron2023llama}
Touvron, H.; Lavril, T.; Izacard, G.; Martinet, X.; Lachaux, M.-A.; Lacroix, T.; Rozi{\`e}re, B.; Goyal, N.; Hambro, E.; Azhar, F.; et~al. 2023.
\newblock LLaMA: Open and Efficient Foundation Language Models.
\newblock \emph{arXiv preprint arXiv:2302.13971}.

\bibitem[{Wang et~al.(2025{\natexlab{a}})Wang, Qiang, Song, Zheng, and Xiong}]{wang2025learning}
Wang, J.; Qiang, W.; Song, Z.; Zheng, C.; and Xiong, H. 2025{\natexlab{a}}.
\newblock Learning to Think: Information-Theoretic Reinforcement Fine-Tuning for LLMs.
\newblock arXiv:2505.10425.

\bibitem[{Wang et~al.(2025{\natexlab{b}})Wang, Yao, Zhou, Hu, Wang, Guan, and Jiang}]{wang2025insightsverificationtrainingverilog}
Wang, N.; Yao, B.; Zhou, J.; Hu, Y.; Wang, X.; Guan, N.; and Jiang, Z. 2025{\natexlab{b}}.
\newblock Insights from Verification: Training a Verilog Generation LLM with Reinforcement Learning with Testbench Feedback.
\newblock arXiv:2504.15804.

\bibitem[{Zhang et~al.(2024)Zhang, Zhoubian, Hu, Yue, Dong, and Tang}]{zhang2024rest}
Zhang, D.; Zhoubian, S.; Hu, Z.; Yue, Y.; Dong, Y.; and Tang, J. 2024.
\newblock Rest-mcts*: Llm self-training via process reward guided tree search.
\newblock \emph{Advances in Neural Information Processing Systems}, 37: 64735--64772.

\bibitem[{Zhang et~al.(2025)Zhang, Hong, Bao, Jiang, Song, Hong, and Xiong}]{zhang2025gvpo}
Zhang, K.; Hong, Y.; Bao, J.; Jiang, H.; Song, Y.; Hong, D.; and Xiong, H. 2025.
\newblock GVPO: Group variance policy optimization for large language model post-training.
\newblock \emph{arXiv preprint arXiv:2504.19599}.

\bibitem[{Zhang and Bowman(2018)}]{zhang2018language}
Zhang, K.~W.; and Bowman, S.~R. 2018.
\newblock Language modeling teaches you more syntax than translation does: Lessons learned through auxiliary task analysis.
\newblock \emph{arXiv preprint arXiv:1809.10040}.

\bibitem[{Zhao et~al.(2025)Zhao, Zhou, Li, Tang, Wang, Hou, Min, Zhang, Zhang, Dong et~al.}]{zhao2023survey}
Zhao, W.~X.; Zhou, K.; Li, J.; Tang, T.; Wang, X.; Hou, Y.; Min, Y.; Zhang, B.; Zhang, J.; Dong, Z.; et~al. 2025.
\newblock A survey of large language models.

\end{thebibliography}

\newpage

\section{Proofs}
\label{sec_app:proof}

\subsection{Proofs of Theorem 1}

We interpret all functions as square-integrable elements of $L^2(\Omega)$, where $\Omega$ is the sample space with probability measure $\mathbb{P}$ induced by $p(X', Y')$.
We define $\Phi$ as a conditional expectation operator:
\begin{equation}
    \Phi f := \mathbb{E}[f \mid q, y_{1:n-1}].
\end{equation}
Its function is to project any function $f \in L^2(\Omega)$ onto the function subspace corresponding to the sub-$\sigma$-algebra spanned by the variables ${q, y_{1:n-1}}$.
Since conditional expectation is an orthogonal projection in $L^2$, this implies: (i) Range: all $\sigma(q, y_{1\:n-1})$-measurable functions with finite second moment. (ii) Kernel: all functions orthogonal to this subspace.
Thus, $\Phi$ is a non-expansive, self-adjoint, idempotent projection operator.
Similarly, $\Psi = \mathrm{Id} - \Phi$ is a projection onto the orthogonal complement.

Let $f = \pi^*(y_0 \mid x)$ and define its projection version as:
\begin{equation}
    f_{\mathrm{proj}} := \Psi f + \pi^*(y_0 \mid q)
\end{equation}
where $\pi^*(y_0 \mid q)$ is the initial estimate under the unconditional output history information.
Our goal is to prove that:
\begin{equation}
    \Delta(f, f_{\mathrm{proj}}) := \mathbb{E}[(Y' - f)^2] - \mathbb{E}[(Y' - f_{\mathrm{proj}})^2] \ge 0.    
\end{equation}
According to the conditional independence structure implied by formula (1) in the original text, under the conditions of given $q$ and $y_{1:n-1}$:
\begin{equation}
    \mathbb{E}[\pi^*(y_0 \mid x) \mid q, y_{1:n-1}] = \pi^*(y_0 \mid q),
\end{equation}
That is $\Phi f = \pi^*(y_0 \mid q)$. 
Combined with $\Psi = \mathrm{Id} - \Phi$, we can get:
\begin{equation}
    f = \Phi f + \Psi f = \pi^*(y_0 \mid q) + \Psi f = f_{\mathrm{proj}}.
\end{equation}
This means that $f$ and $f_{\mathrm{proj}}$ are actually equal. This yields $\Delta = 0$ under ideal conditions. However, this result becomes non-trivial in practice when $\pi^*$ deviates from the perfect Bayesian estimator due to finite data or model approximation. In order to obtain a non-trivial generalized inequality, we further use geometric methods to expand the error term and reveal the structural return.

Next, let $Y' := y_0$ be the prediction target, $f := \pi^*(y_0 \mid x)$ be the original predictor, and $f_{\mathrm{proj}} := \Psi f + \pi^*(y_0 \mid q)$ be the projection form.
Consider the squared error:
\begin{equation}
    \Delta(f, f_{\mathrm{proj}}) = \|Y' - f\|^2 - \|Y' - f_{\mathrm{proj}}\|^2.
\end{equation}
Since $f = f_{\mathrm{proj}} + (\Phi f - \pi^*(y_0 \mid q))$, we write it as $Y' - f = Y' - f_{\mathrm{proj}} - (\Phi f - \pi^*(y_0 \mid q))$. Thus, we have:
\begin{equation}
\begin{split}
        \|Y' - f\|^2 = &\|Y' - f_{\mathrm{proj}}\|^2 + \|\Phi f - \pi^*(y_0 \mid q)\|^2\\
        &+ 2 \langle Y' - f_{\mathrm{proj}}, \Phi f - \pi^*(y_0 \mid q) \rangle.
\end{split}
\end{equation}
However, since $\Phi f - \pi^*(y_0 \mid q) = 0$, the cross term disappears, so $\Delta(f, f_{\mathrm{proj}}) = \|\Phi f - \pi^*(y_0 \mid q)\|^2 = 0$. If the model $\pi^*$ is not a perfect Bayesian optimal estimator, or disturbances are introduced during training, then $\Phi f \ne \pi^*(y_0 \mid q)$, and we have:
\begin{equation}
    \Delta(f, f_{\mathrm{proj}}) = \|\Phi f - \pi^*(y_0 \mid q)\|^2 \ge 0.
\end{equation}
Therefore, we have $\Delta(f, f_{\mathrm{proj}}) \ge 0$, completing the proofs.

\subsection{Proofs of Corollary 2}

We aim to show that the following inequality holds under the same structural assumptions as Theorem 1:
\begin{equation}
\Delta\left(\pi^*(y_0 \mid q),\; \Psi \cdot \pi^*(y_0 \mid x) + \pi^*(y_0 \mid q)\right) \ge 0.
\end{equation}
To this end, we interpret all functions as square-integrable elements of $L^2(\Omega)$, where $\Omega$ is the sample space equipped with the probability measure $\mathbb{P}$ induced by the joint distribution $p(X', Y')$ of input-output pairs.

We define $\Phi$ as the conditional expectation operator $\Phi f := \mathbb{E}[f \mid q, y_{1\:n-1}]$, which projects any $f \in L^2(\Omega)$ onto the subspace of functions measurable with respect to the $\sigma$-algebra generated by ${q, y_{1\:n-1}}$.

By the standard properties of conditional expectation in $L^2$, this operator is an orthogonal projection, meaning it is self-adjoint, idempotent, and non-expansive, and its range and null space are orthogonal complements.

We further define $\Psi := \mathrm{Id} - \Phi$ as the projection onto the orthogonal complement of $\mathrm{Range}(\Phi)$.

Let $f := \pi^*(y_0 \mid x)$ be the Bayes-optimal prediction based on the full input $x = {q, y_1, \dots, y_{n-1}}$, and let $b := \pi^*(y_0 \mid q)$ denote the model’s prediction when only the query $q$ is observed.
We then define the causally projected prediction as $f_{\mathrm{proj}} := \Psi f + b$, which augments the baseline estimate $b$ with the residual component of $f$ that is orthogonal to $\sigma(q, y_{1\:n-1})$.

Our goal is to evaluate the difference in squared prediction error between $b$ and $f_{\mathrm{proj}}$, defined by
\begin{equation}
\Delta(b, f_{\mathrm{proj}}) := \mathbb{E}[(Y' - b)^2] - \mathbb{E}[(Y' - f_{\mathrm{proj}})^2].
\end{equation}

To compute this quantity, we note that since $f_{\mathrm{proj}} = \Psi f + b$, we have
\begin{equation}
Y' - b = (Y' - f_{\mathrm{proj}}) + \Psi f,
\end{equation}
and thus, by expanding the square norm, it follows that
\begin{equation}
\begin{split}
|Y' - b|^2 &= |Y' - f_{\mathrm{proj}} + \Psi f|^2 \\
&= |Y' - f_{\mathrm{proj}}|^2 + |\Psi f|^2 + 2 \langle Y' - f_{\mathrm{proj}}, \Psi f \rangle.
\end{split}
\end{equation}

Subtracting $|Y' - f_{\mathrm{proj}}|^2$ from both sides yields the regret difference
\begin{equation}
\Delta(b, f_{\mathrm{proj}}) = \|\Psi f\|^2 + 2 \langle Y' - f_{\mathrm{proj}}, \Psi f \rangle.
\end{equation}

Since $f_{\mathrm{proj}} = \Psi f + b$, we further have $Y' - f_{\mathrm{proj}} = Y' - b - \Psi f$, and so
\begin{equation}
\langle Y' - f_{\mathrm{proj}}, \Psi f \rangle = \langle Y' - b - \Psi f, \Psi f \rangle = \langle Y' - b, \Psi f \rangle - \|\Psi f\|^2.
\end{equation}

Substituting back, we obtain:
\begin{equation}
\begin{split}
\Delta(b, f_{\mathrm{proj}}) &= \|\Psi f\|^2 + 2 (\langle Y' - b, \Psi f \rangle - \|\Psi f\|^2) \\
&= - \|\Psi f\|^2 + 2 \langle Y' - b, \Psi f \rangle.
\end{split}
\end{equation}
Now, we invoke the conditional independence result established in Equation\~(1), which implies that $\mathbb{E}[Y'| q, y_{1\:n-1}] = \pi^*(y_0 \mid q) = b$, and hence
\begin{equation}
\mathbb{E}[Y' - b \mid q, y_{1:n-1}] = 0.
\end{equation}
As a result, the inner product $\langle Y' - b, \Psi f \rangle = \mathbb{E}[(Y' - b)\Psi f] = 0$, since $\Psi f$ is orthogonal to the $\sigma(q, y_{1\:n-1})$-measurable subspace to which $Y' - b$ belongs in expectation.

Therefore, we conclude that
\begin{equation}
\Delta(b, f_{\mathrm{proj}}) = \|\Psi f\|^2 \ge 0,
\end{equation}
which completes the proof.

\section{Pseudo-code of GCPO}
\label{sec_app:pseudo-code}
The pseudo-code of our proposed GCPO is shown in Algorithm \ref{alg:algorithm}.

\begin{algorithm}[h]
\caption{Pseudo-Code of GCPO}
\label{alg:algorithm}
\begin{algorithmic}[1]
\REQUIRE Initial policy $\pi_\theta$; prompt distribution $\mathcal{D}$; hyperparameters  $\alpha$, $\beta$, and $\kappa$
\FOR{step $= 1$ \TO $n$}
    \STATE Sample a batch $D_b$ from $\mathcal{D}$
    \STATE Set old policy $\pi_{\theta_{\rm old}} \leftarrow \pi_\theta$
    \FOR{each query $q \in D_b$}
        \STATE Sample group $\{y_0, y_1, \cdots, y_{n-1}\} \sim \pi_{\theta_{\rm old}}(\cdot|q)$
        \STATE Sample group $\{y_{n,i}\}_{i=0}^{n-1} \sim \pi_{\theta_{\rm old}}(\cdot|q,\{y_i\}_{i=0}^{n-1})$
        \FOR{each $y_i$}
            \STATE Construct $x_i = \{q, y_0, \ldots, y_{n-1}\} \setminus \{y_i\}$
            \STATE Compute reward $r_i$ and advantage $A_i$ via Eq.(8)
            \STATE Compute causal factor $\Upsilon_i$ via Eq.(11)
            \STATE Obtain the relative advantage $B_i = A_i \cdot \Upsilon_i$
            \STATE Compute $D_{\rm KL}(\pi_\theta \parallel \pi_{\rm ref})$
            \STATE Construct $\pi'_{\rm ref}$ and compute $D_{\rm KL}(\pi_\theta \parallel \pi'_{\rm ref})$
        \ENDFOR
    \ENDFOR
    \STATE Update $\pi_\theta$ via the GCPO objective in Eq.(10)
\ENDFOR
\RETURN $\pi_\theta$
\end{algorithmic}
\end{algorithm}


\section{Benchmark Datasets}
\label{sec_app:benchmark}
This section provides a brief overview of the datasets used in our experiments. Broadly, the benchmarks fall into two categories:
(i) reasoning tasks for mathematical derivation, including AIME24-25, AMC, MATH500 \cite{hendrycks2021measuring}, MinervaMATH \cite{lewkowycz2022solving}; and
(ii) reasoning tasks for code generation, i.e., HumanEval \cite{chen2021evaluating}.
The composition and characteristics of each benchmark are summarized as follows.

\textbf{AIME24-25} contains 30 fill-in-the-blank questions drawn from the 2024 and 2025 American Invitational Mathematics Examinations (15 questions per year). These problems are generally more challenging than those in AMC, covering number theory, combinatorics, geometry, and algebra.

\textbf{AMC} includes 975 multiple-choice questions from 39 AMC competitions, with 25 questions each for AMC10 (targeted at students up to 10th grade) and AMC12 (up to 12th grade). The problems range from basic algebra and geometry to introductory topics in probability and combinatorics, offering a diverse set of tasks for evaluating LLM reasoning.

\textbf{MATH500} is a subset of 500 problems randomly sampled from the full MATH dataset. It spans seven mathematical domains, including prealgebra, algebra, number theory, geometry, intermediate algebra, and precalculus. Each problem is accompanied by a step-by-step solution and a difficulty label ranging from 1 to 5, allowing for fine-grained assessment of mathematical reasoning performance.

\textbf{MinervaMATH} contains 12,500 high school-level competition-style math problems. Each question includes detailed solution steps and covers a broad curriculum from prealgebra to precalculus.

\textbf{HumanEval} consists of 164 Python programming tasks designed to evaluate the correctness of code generated by language models. Each task includes a function signature and a natural language description, requiring the model to produce a working implementation. Evaluation is based on the $Pass@k$ metric, which measures the proportion of times the generated code passes all test cases within $k$ attempts.

\section{Implementation Details}
\label{sec_app:implementation}
Our implementation is based on the TRL codebases, with custom modifications. For model initialization, we directly load base models from Hugging Face, including DeepScaleR-1.5B-Preview, DeepSeek-R1-Distill-Qwen-1.5B, DeepSeek-R1-Distill-Qwen-7B, and Qwen2-7B-Instruct. Unless otherwise stated, we follow the official evaluation protocols of each benchmark and report maj@4 scores across different models.
For certain mathematical tasks, DeepScaleR-1.5B-Preview is first fine-tuned on a dataset of 40,000 math problems and solutions, and then further fine-tuned on 919 AIME problems from 1989 to 2023. For DeepSeek-R1-Distill-Qwen-1.5B, we fine-tune using a random subset of 4,000 problem-solution pairs sampled from the NuminaMath dataset. All fine-tuning is performed with a maximum token budget of 16,384 tokens, which also serves as the evaluation constraint.
The training configuration is as follows: we set the learning rate to $1.0 \times 10^{-6}$, employ a cosine learning rate scheduler with a warm-up ratio of 0.1, and use a batch size of 256. The maximum prompt length is 4,096 tokens, and the maximum generation length is 16,384 tokens. Each model is trained for up to 10 epochs, with early stopping typically at 1 epoch.
We enable vLLM acceleration by setting the `use\_vllm' flag to True, with GPU memory utilization capped at 80\%. Mixed-precision training is employed using BF16. The regularization coefficients $\alpha$ and $\kappa$ are set to 2 and 0.06, respectively, based on grid search results.

All experiments were conducted on A100 clusters, i.e., a high-performance GPU cluster consisting of multiple interconnected nodes, each equipped with 8× NVIDIA A100-SXM4 40GB GPUs and an AMD EPYC 7742 64-core CPU, with 512 GB RAM per node. The cluster supports multi-node distributed training via NCCL and InfiniBand, enabling efficient fine-tuning of large-scale models such as 7B GCPO, with DeepSpeed 0.13.1 plus ZeRO-3 for large model optimization. All training is performed with mixed-precision (bf16) under a Slurm-based job scheduling environment.

\begin{figure*}[htpb]
    \centering
    \centering
    \subfigure[Example 1: 4-digit numbers divisible by 5]{
        \includegraphics[width=0.48\linewidth]{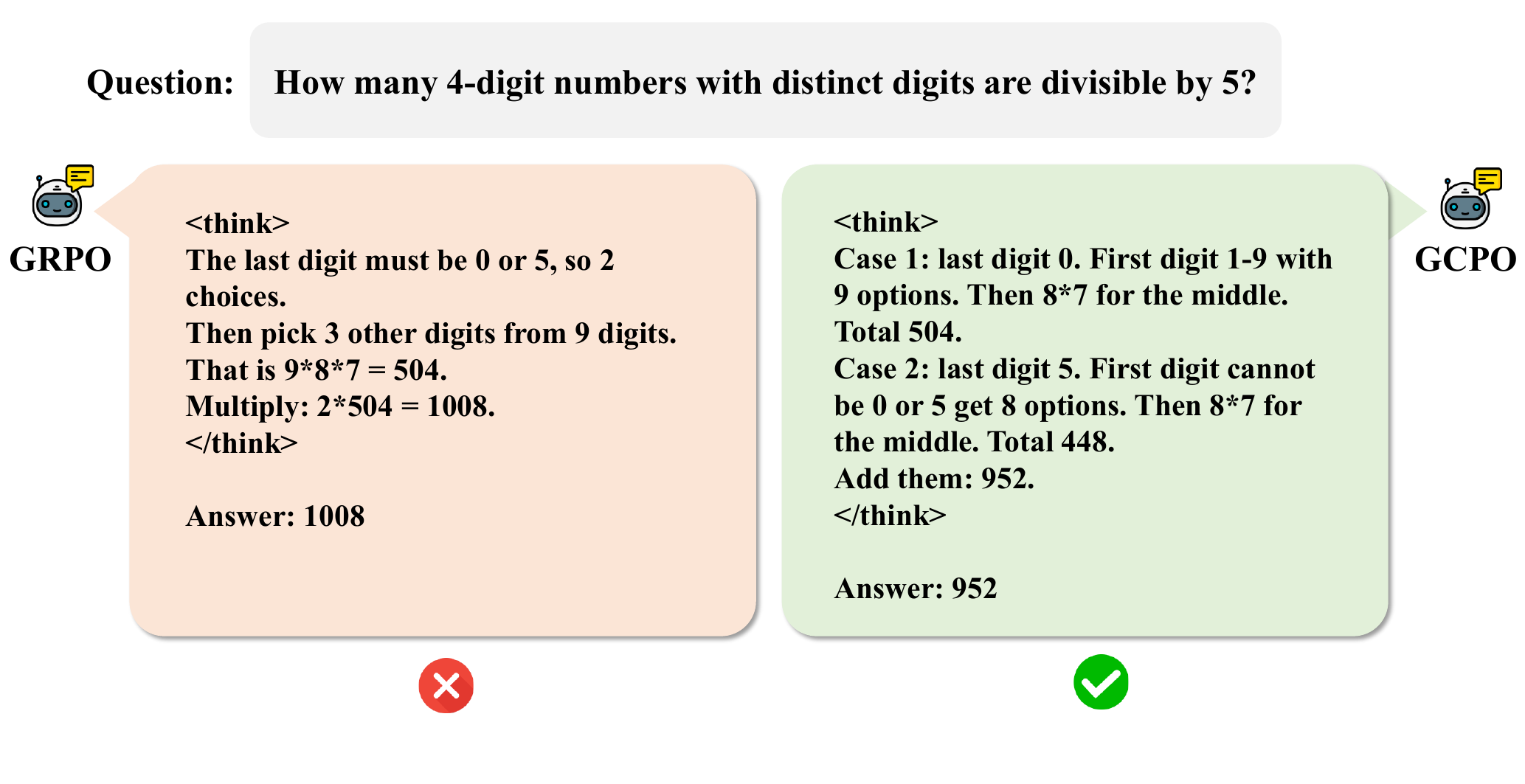}}
        \hfill
    \subfigure[Example 2: Expected number of consecutive heads]{
        \includegraphics[width=0.48\linewidth]{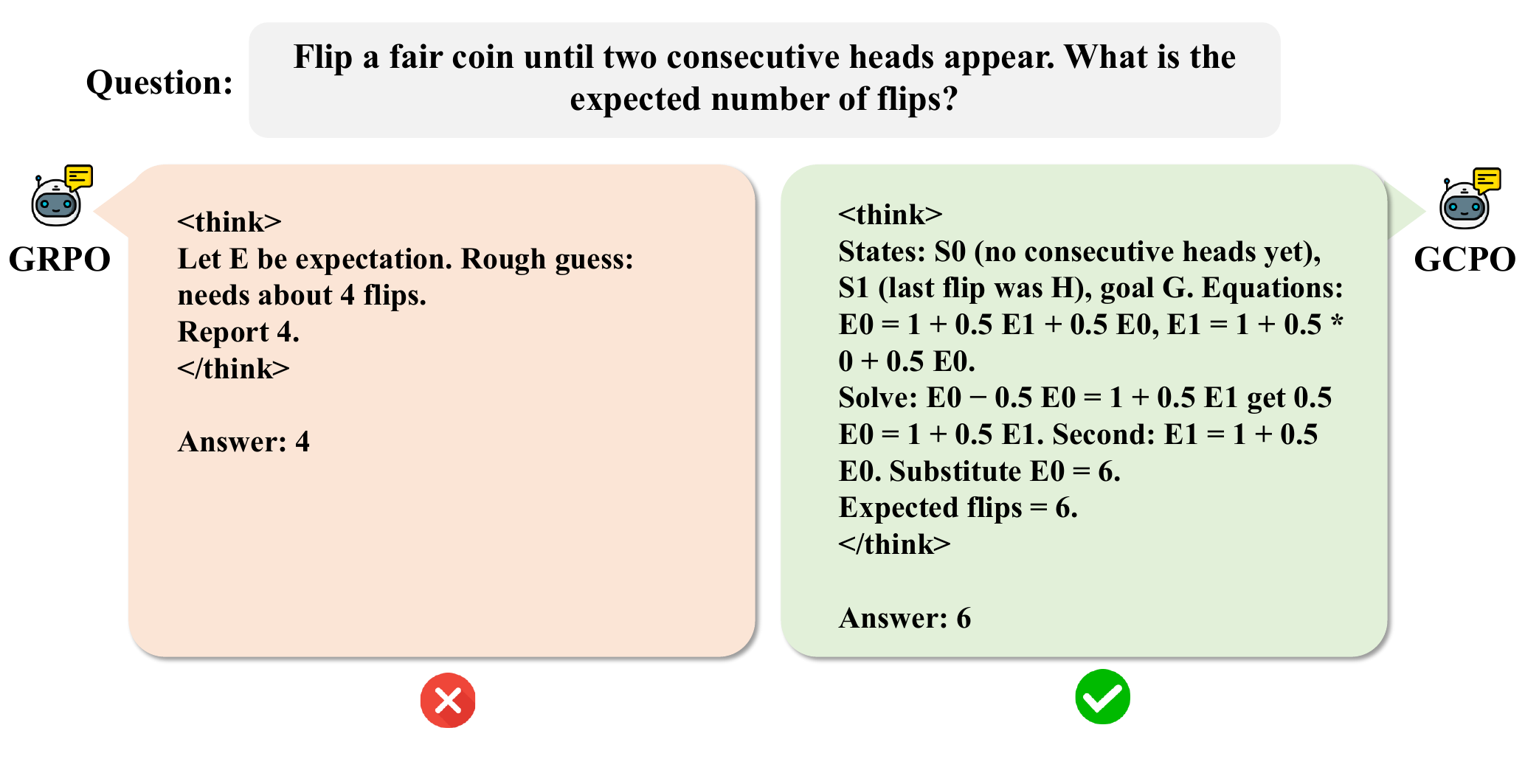}}
    \subfigure[Example 3: Triangle with integer sides and area 60]{
        \includegraphics[width=0.48\linewidth]{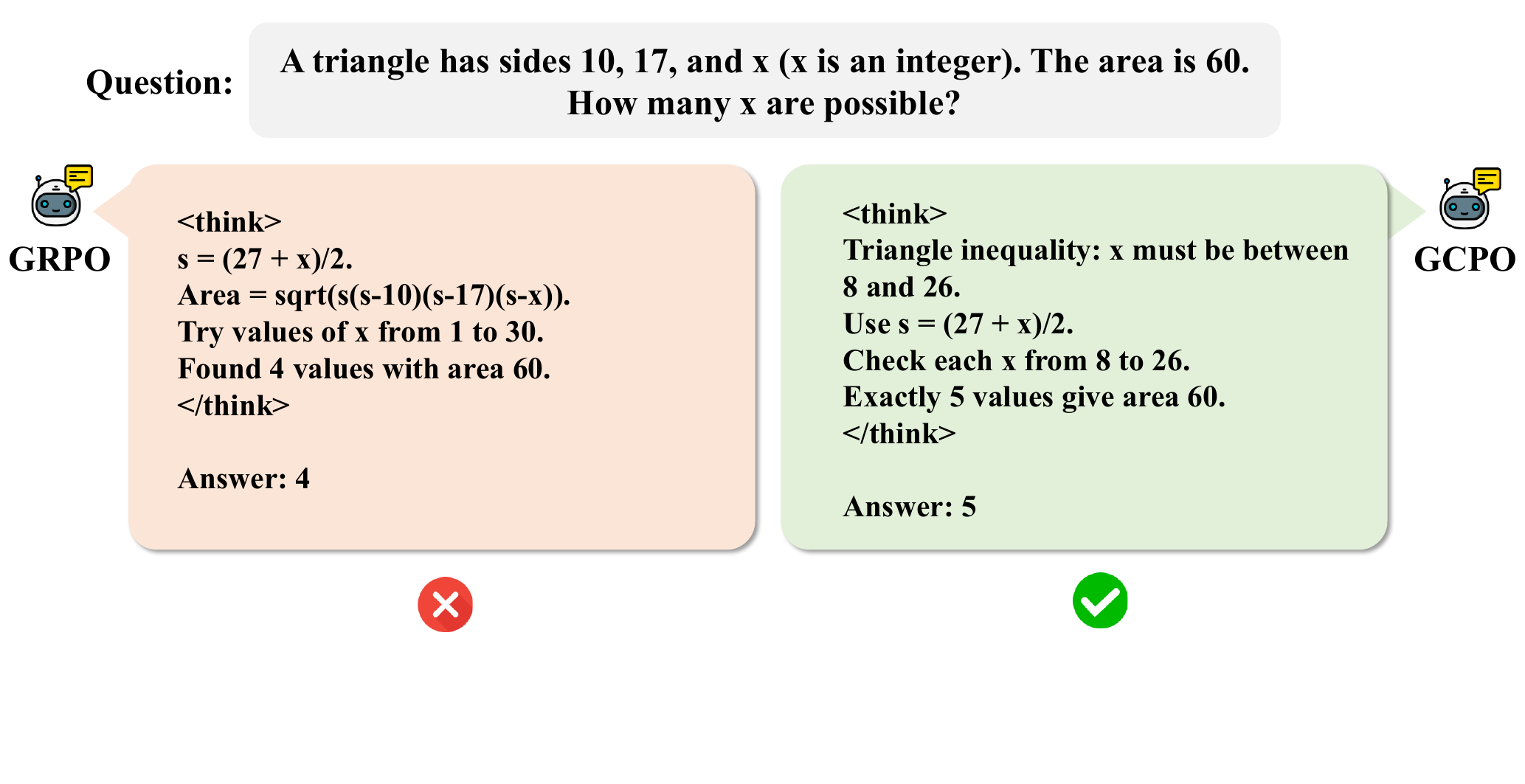}}
    \subfigure[Example 4: Numbers that are both perfect squares and cubes]{
        \includegraphics[width=0.48\linewidth]{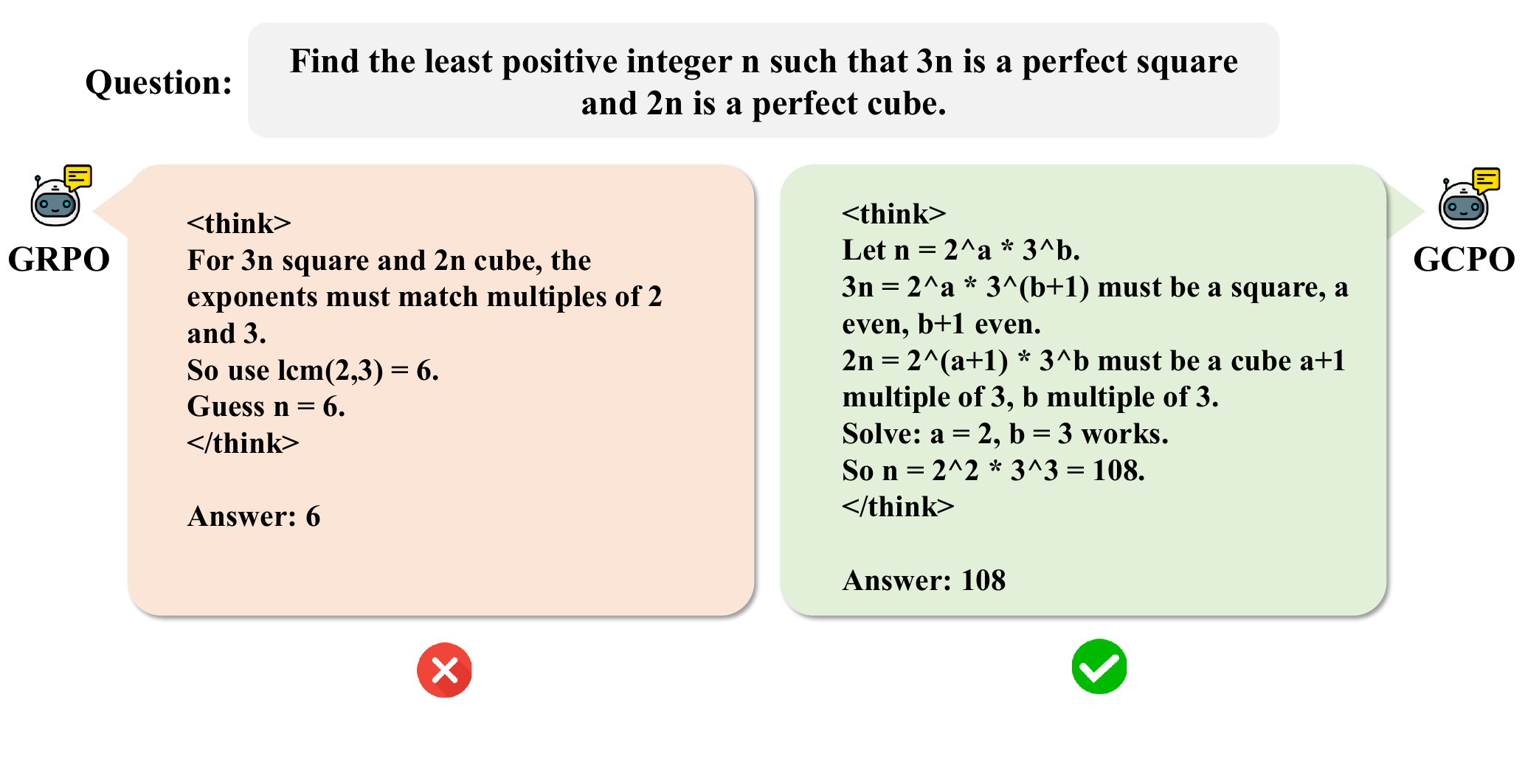}}
    \subfigure[Example 5: Probability of drawing balls]{
        \includegraphics[width=0.48\linewidth]{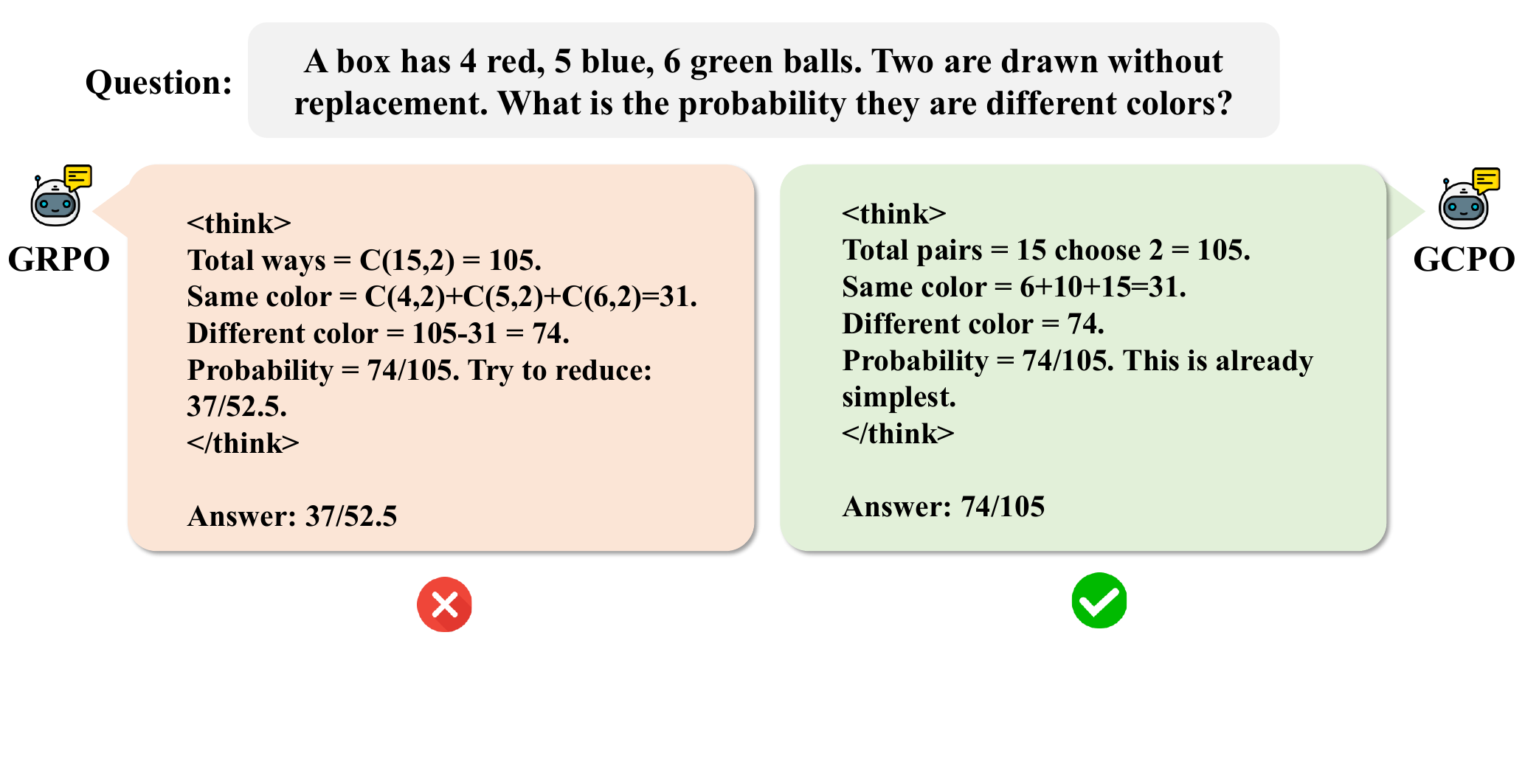}}
    \subfigure[Example 6: Divisibility filtering]{
        \includegraphics[width=0.48\linewidth]{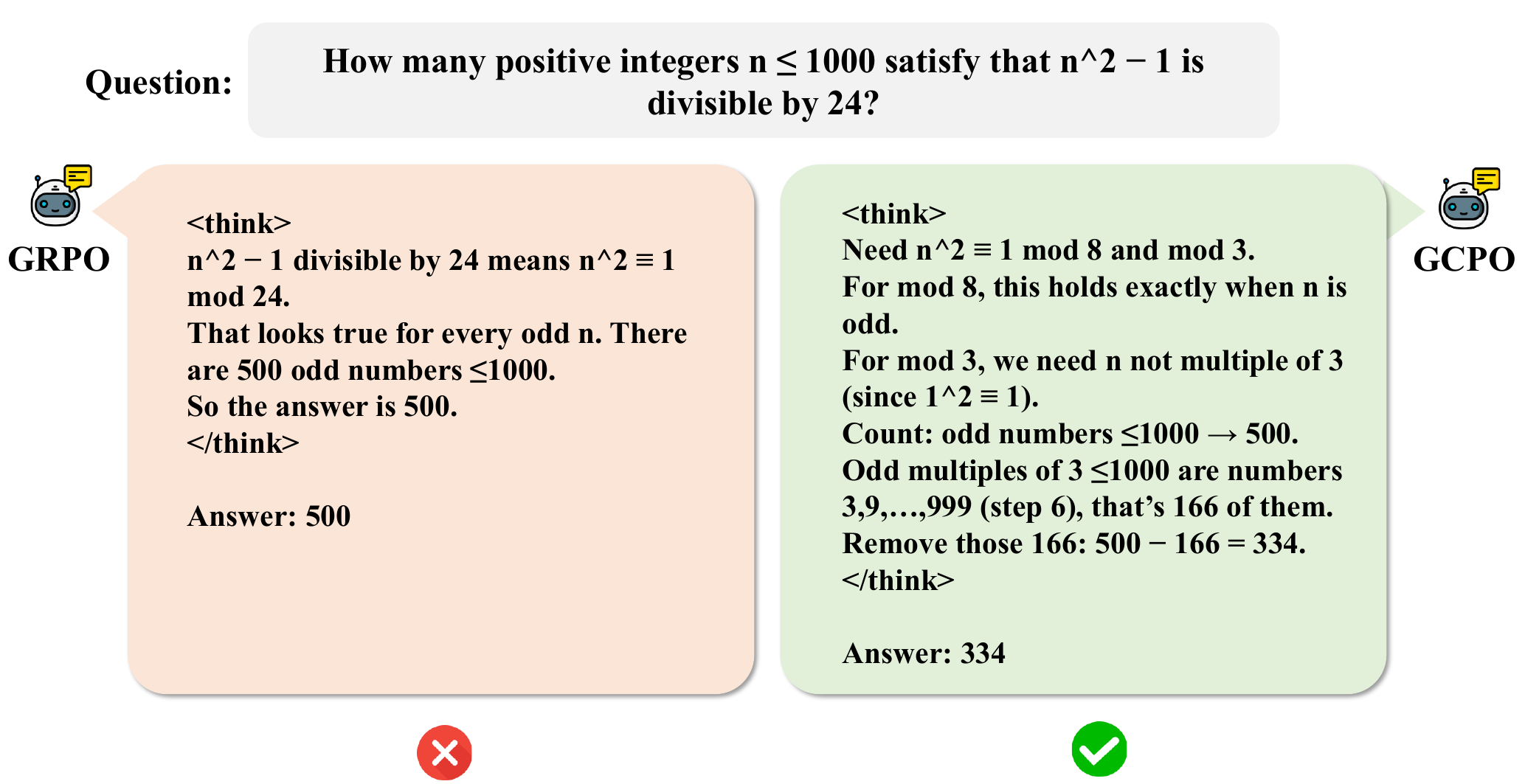}}
    \subfigure[Example 7: Counting surjective mappings]{
        \includegraphics[width=0.48\linewidth]{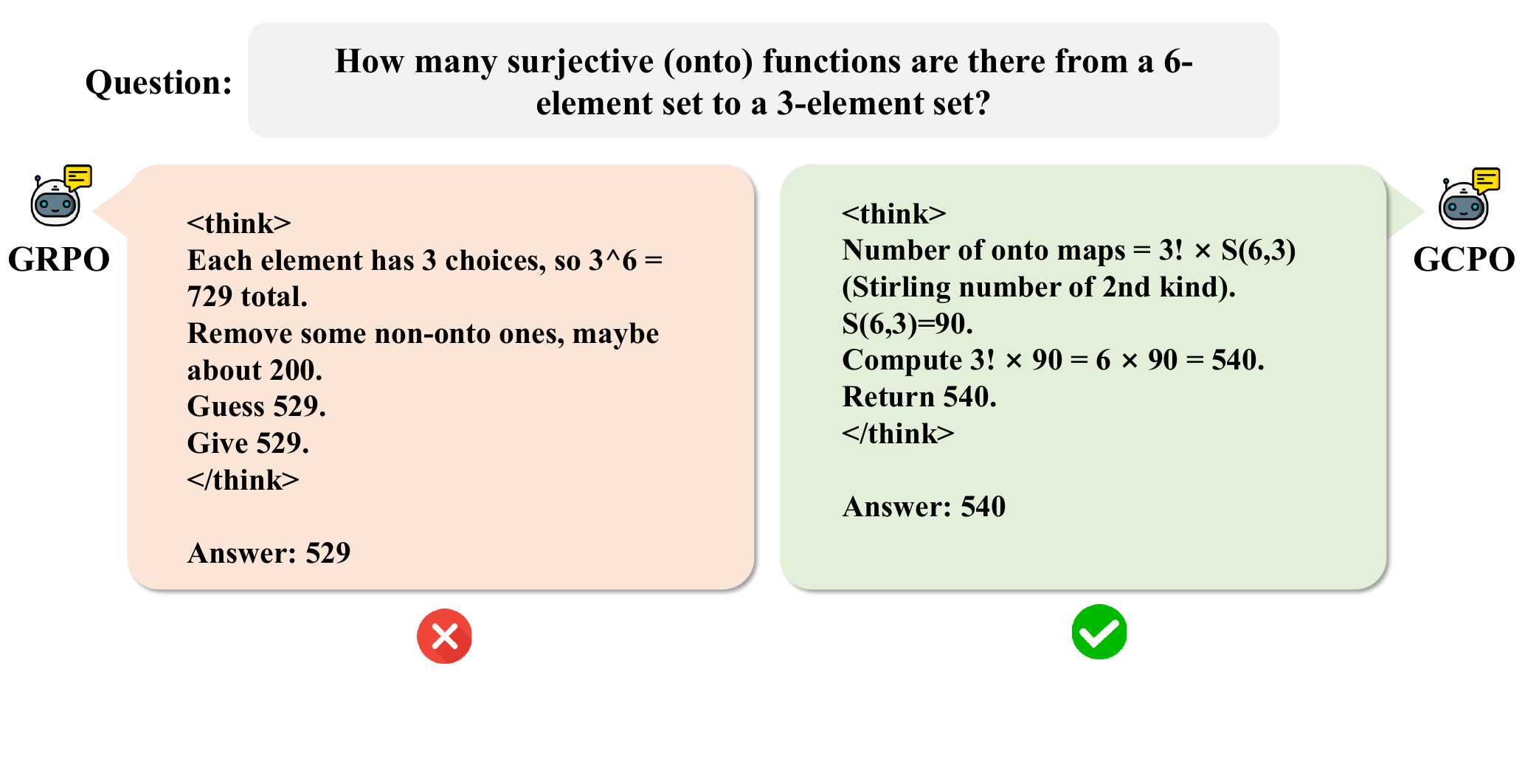}}
    \subfigure[Example 8: Lattice paths that stay within bounds]{
        \includegraphics[width=0.48\linewidth]{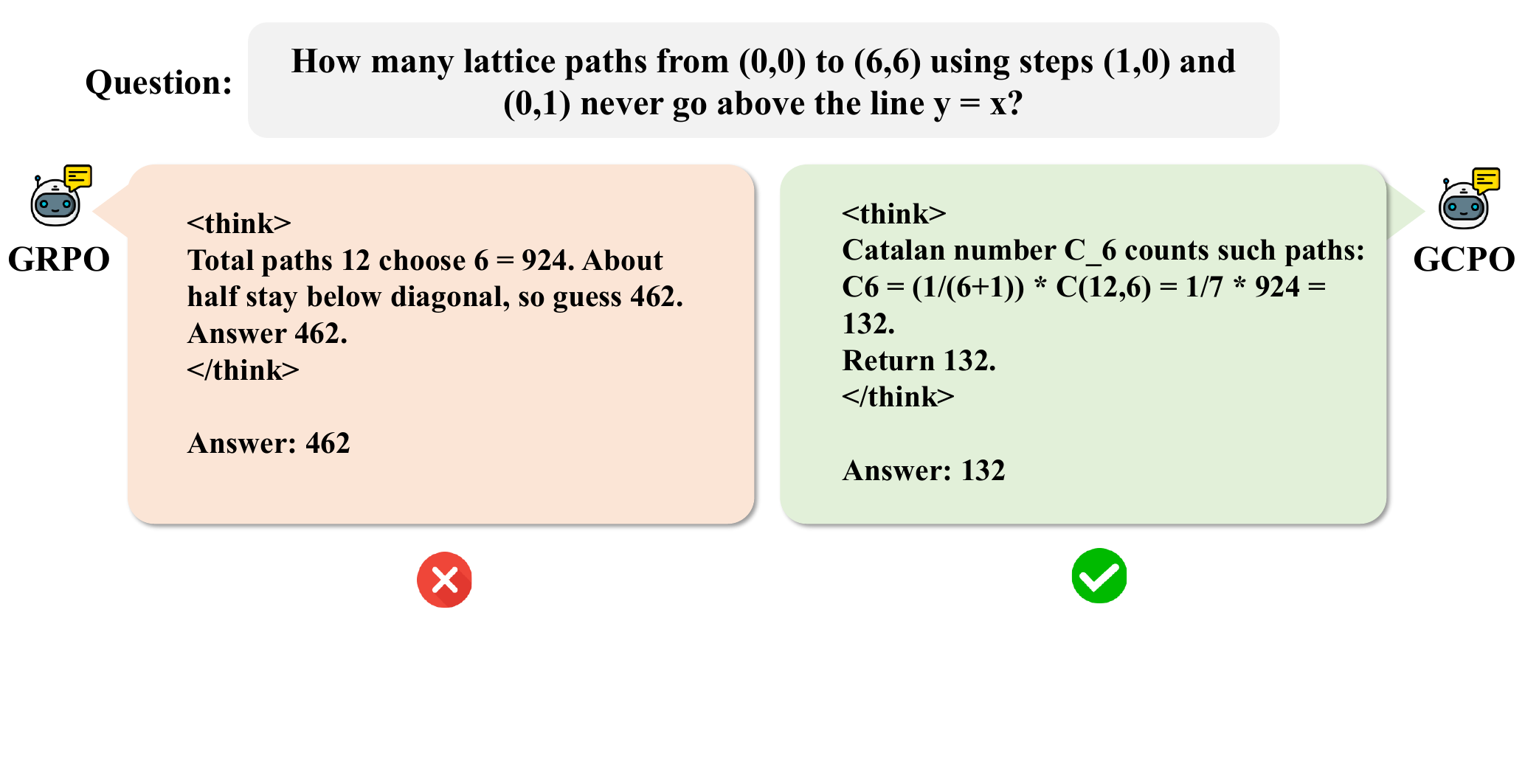}}
    \caption{Qualitative analysis of side-by-side rollouts. For each question, we show a GRPO (left) and a GCPO chain (right). GRPO lacks cross-component causal coordination, leading to systematic errors (e.g., missing modular constraints, miscounting Catalan paths). GCPO organizes multi-aspect constraints within identical step budgets, resolves the errors, and returns the correct answers across topics ranging from number theory and geometry to probability, combinatorics, and lattice-path counting.}
    \label{fig:app_qua}
\end{figure*}

\begin{figure}[t]
    \centering
    \centering
    \subfigure[Evaluation on GRPO]{
        \includegraphics[width=0.48\linewidth]{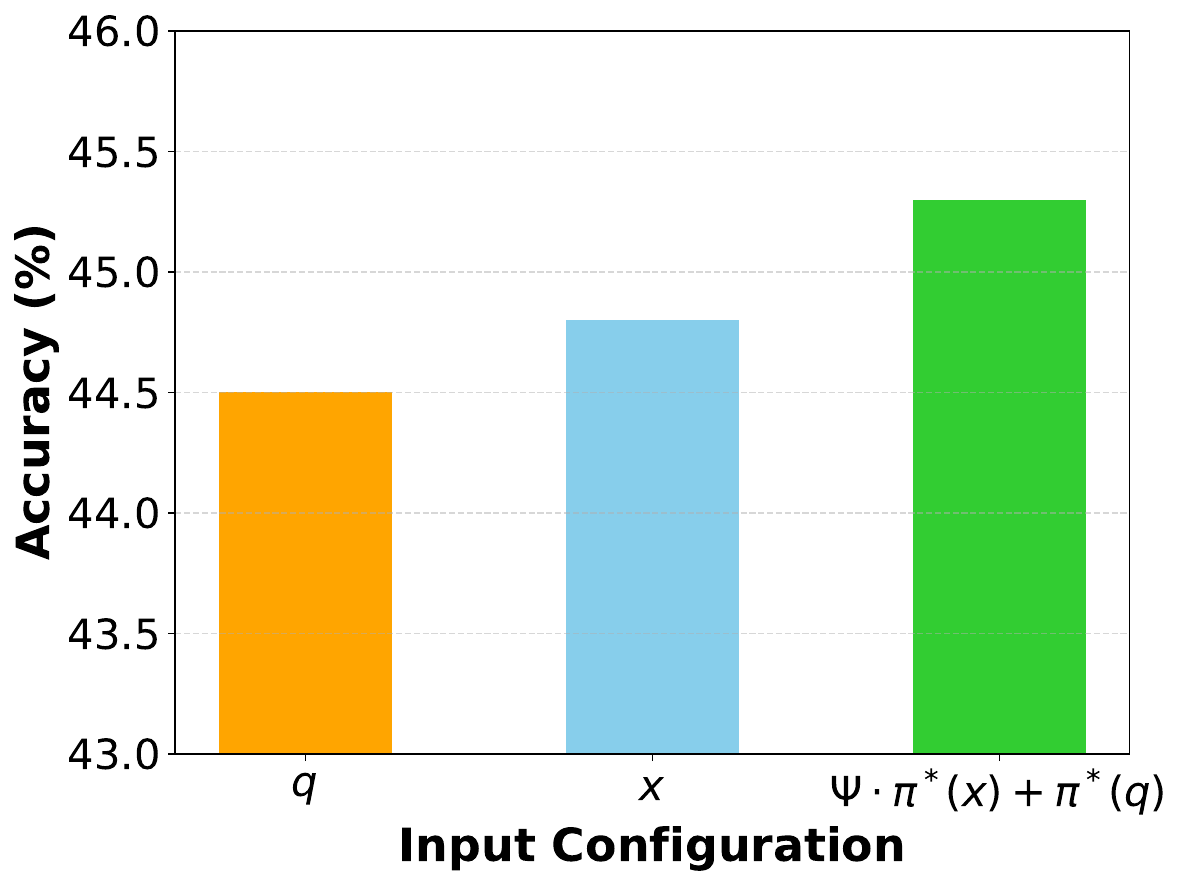}}
        \hfill
    \subfigure[Evaluation on GCPO]{
        \includegraphics[width=0.48\linewidth]{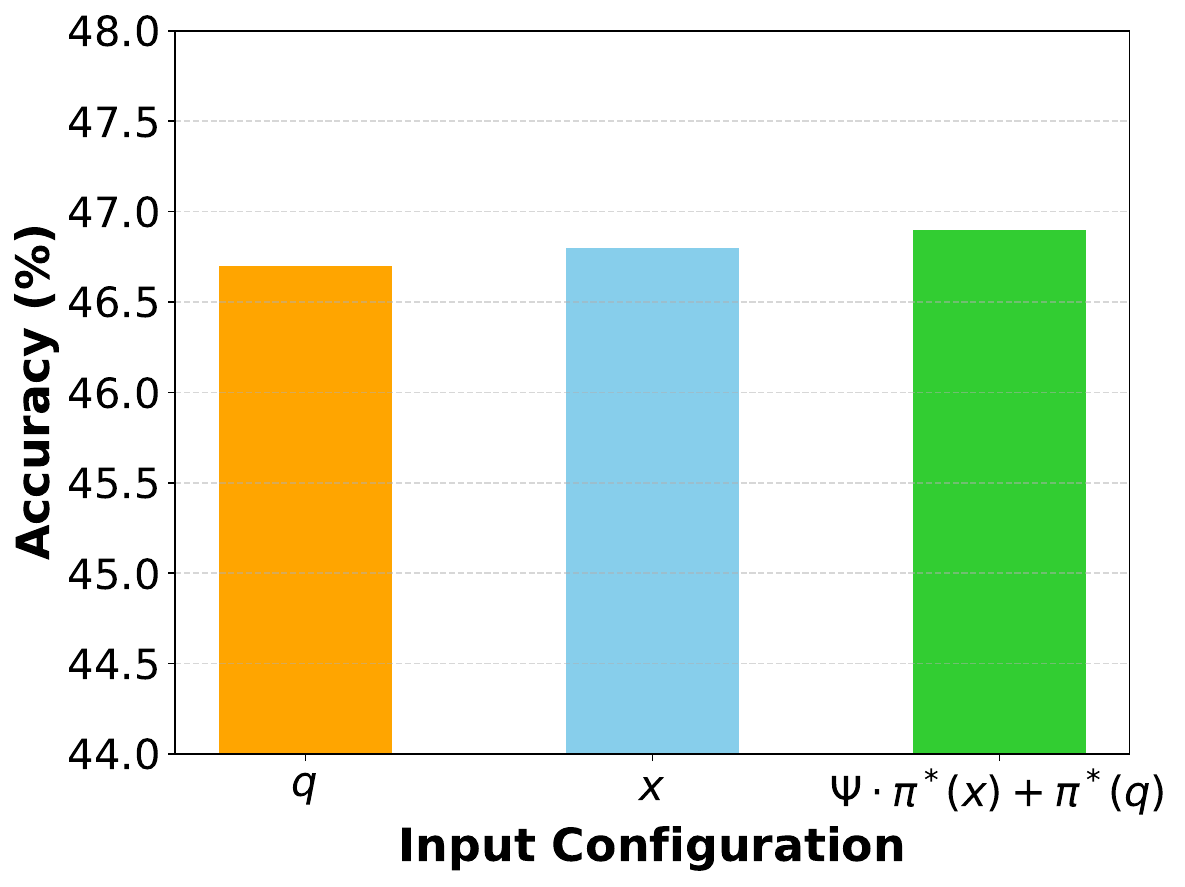}}
    \caption{Ablation results under GRPO and GCPO frameworks using three input configurations.}
    \label{fig:app_ablation}
\end{figure}

\begin{table*}[t]
  \centering
  \small
    \begin{tabular}{l|c|c|c|c|c|c}
      \toprule
      Base model + Method & AIME 2024 & AIME 2025 & AMC 2023 & MATH500 & MinervaMATH & \textbf{Avg.} \\
      \midrule
      \!\textbf{DeepScaleR‑1.5B‑Preview} & 42.8 & 36.7 & 83.0 & 85.2 & 24.6 & 54.5 \\
      \!~~+GRPO & 44.5 & 39.3 & 81.5 & 84.9 & 24.7 & 55.0 \\
      \!~~+GRPO + MCTS & 45.3 & 40.2 & 82.4 & 85.4 & 25.1 & 55.7 \\
      \!~~+GVPO & 46.1 & 39.7 & 83.6 & 85.7 & 25.3 & 56.1 \\
      \!~~+GVPO + MCTS & 46.8 & 40.5 & 83.9 & 86.2 & 25.6 & 56.6 \\
      \!~~+Dr.GRPO & 45.8 & 39.6 & 82.1 & 85.3 & 25.1 & 55.6 \\
      \!~~+Dr.GRPO + MCTS & 46.5 & 40.4 & 82.9 & 85.9 & 25.5 & 56.2 \\
      \!~~+GCPO & 46.7 & 40.3 & 84.1 & 86.3 & 25.9 & 56.8 \\
      \!~~+GCPO + MCTS & 47.5 & 41.2 & 84.6 & 86.9 & 26.3 & 57.3 \\
      \bottomrule
    \end{tabular}
  \caption{Pass@1 performance on various math reasoning benchmarks.}
  \label{tab:ex_app_comparison}
\end{table*}

\begin{figure}[t]
    \centering
    \includegraphics[width=\linewidth]{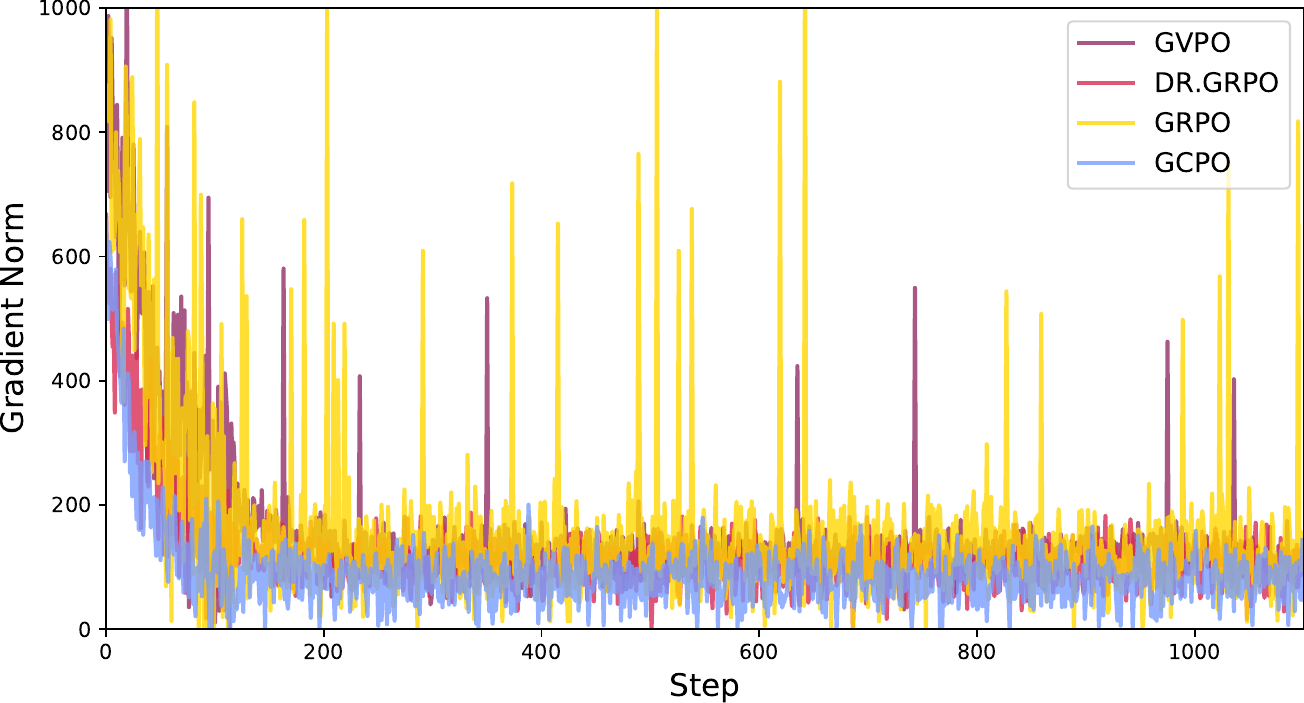}
    \caption{The norm of the gradient during training.}
    \label{fig:norm}
\end{figure}

\begin{figure}[t]
    \centering
    \includegraphics[width=\linewidth]{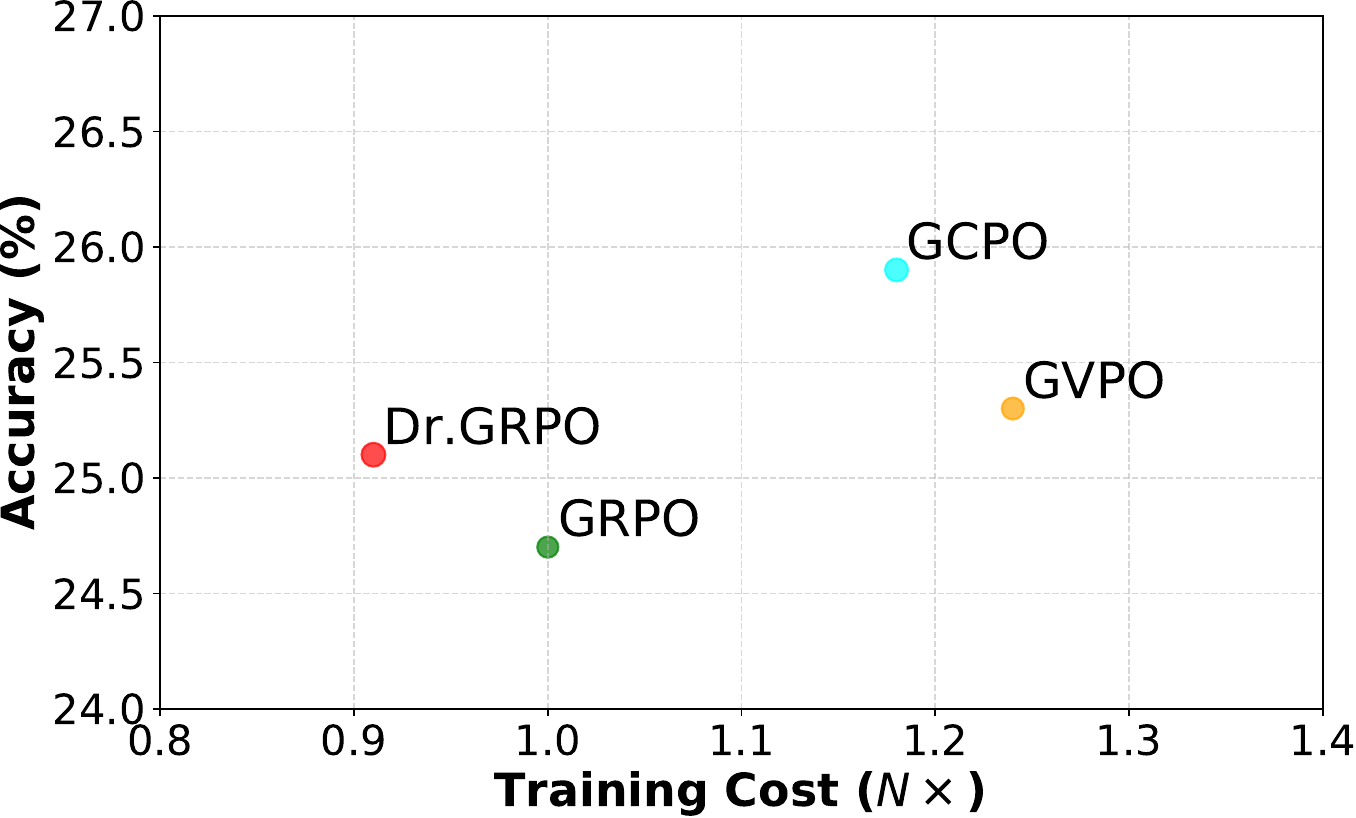}
    \caption{Trade-off performance of different methods.}
    \label{fig:app_cost}
\end{figure}

\section{Additional Experiments and Discussion}
\label{sec_app:experiments}

\subsection{Full Results of Comparison}

We conduct comprehensive experiments across a diverse set of reasoning benchmarks in the main text, including mathematical tasks (AIME24–25, AMC, MATH500, MinervaMATH) and code generation (HumanEval). The experiments are carried out using several open-source base models such as DeepScaleR-1.5B-Preview, DeepSeek-R1-Distill-Qwen-1.5B/7B, and Qwen2-7B-Instruct. GCPO is compared against a range of strong baselines, including GRPO, GVPO, ReST-MCTS, and Dr.GRPO.
The results in the main text show that GCPO consistently outperforms all baselines across benchmarks. 

To provide a more comprehensive evaluation, in this section, we conduct comparison experiments with the SOTA test-time methods, e.g., MCTS, to explore whether reasoning effective fine-tuning avoids the need for test-time compute, or further gains can be achieved.
Specifically, we directly introduce MCTS on models trained based on GRPO, Dr.GRPO, and GCPO. Then, we record the accuracy of these models before and after introducing the corresponding methods. The results are shown in Table \ref{tab:ex_app_comparison}. From the results, we can observe that GCPO still achieves the best performance. In addition, while these test-time methods provide certain performance improvements, they still fail to surpass GCPO even when combined with RL-based training baselines. This demonstrates the irreplaceability of the causal module introduced by GCPO, i.e., it uncovers inter-group relationships that were overlooked by previous work and guides the model to learn them for further reasoning optimization.

\subsection{Training Stability}
Given the substantial computational cost of training LLMs, maintaining stable training dynamics is crucial to ensure convergence efficiency and avoid catastrophic failures during optimization. Unstable updates can lead to divergent behavior, increased variance in model performance, and wasted computational resources. To quantitatively assess training stability, we adopt the gradient norm as a proxy for policy variance, following standard practice in reinforcement learning. A stable gradient norm suggests consistent updates to the model parameters, whereas large fluctuations may indicate unstable or overly aggressive policy shifts.
As shown in Figure \ref{fig:norm}, GCPO exhibits the most stable training behavior among all compared methods, with its gradient norm remaining nearly constant across training iterations. This indicates that the policy updates are well-regulated, likely due to the regularization effect introduced by the causal term in GCPO. In contrast, GRPO suffers from pronounced oscillations in gradient norm, reflecting unstable dynamics that may hinder reliable convergence. These results underscore the importance of incorporating causal constraints into the learning process.

\subsection{Computational Overhead Analysis}
To assess the training efficiency of different methods, we normalize the computational cost of GRPO to $1\times$ as a baseline, and compare all other methods accordingly. The cost is measured in terms of total GPU hours under matched training schedules and hardware settings.
The results are shown in Figure \ref{fig:app_cost}.
We find that GCPO incurs a modest increase in training cost (~$1.18\times$) due to the incorporation of a KL-regularized causal objective. Despite this slight overhead, GCPO consistently outperforms all baselines across benchmarks, yielding the highest gains in reasoning accuracy and robustness. This indicates a favorable trade-off between cost and performance.
In summary, GCPO offers a great balance between efficiency and effectiveness, delivering superior results with only a modest increase in training overhead compared to the GRPO baseline.

\subsection{More Ablation Studies}
In the ablation study of the main text, we verify that both terms are crucial to the observed improvements. Further, sensitivity analysis identifies optimal hyperparameters at $\alpha = 2$ and $\beta = 0.06$, with performance remaining stable in a reasonable range. These results demonstrate that GCPO is robust and effective across reasoning tasks of varying complexity and modality. 

In this subsection, we further conduct a ablation study to examine the predictive effects of different input configurations, i.e., the original query $q$, the input $x = {q, y_1, \dots, y_{n-1}}$, and the causally projected input $\Psi \cdot \pi^*(x) + \pi^*(q)$ (as mentioned in Section Causal Analysis).
The experiment is conducted under both GRPO and GCPO frameworks by feeding the three different input configurations into the trained models and evaluating their performance. 
The results are shown in Figure \ref{fig:app_ablation}.
From the results, we can observe that under GRPO, prediction quality varies significantly: $q$ alone leads to the poorest performance, $x$ provides moderate improvement, while the causally projected input achieves the best results. This confirms the benefit of leveraging collider-aware representations. In contrast, GCPO yields consistently high and stable performance across all inputs, indicating that its causal objective effectively internalizes the relevant dependencies, making it robust to input variation.

\subsection{Qualitative Analysis}
To better illustrate the impact of GCPO on model behavior, we conduct a qualitative analysis comparing outputs before and after applying GCPO. Specifically, we randomly sample a set of math reasoning problems from the benchmark datasets. For each problem, we visualize the outputs generated by models trained with GRPO and GCPO, respectively.
We observe that, for certain complex reasoning tasks, the GCPO-trained model is able to explore multiple aspects of the problem, reflect on intermediate steps from different perspectives, and ultimately arrive at the correct solution. In contrast, the GRPO-trained model often follows a relatively rigid line of reasoning, which can lead to errors.

Figure \ref{fig:app_qua} presents representative examples. 
These qualitative results suggest that GCPO enables the model to capture inter-path relationships and reason more effectively by integrating diverse viewpoints during inference.
For example, in Figure \ref{fig:app_qua}(b), GRPO offers a heuristic guess without modeling the underlying process. GCPO distinguishes three Markov states, i.e., no previous head, one previous head, and termination, and sets up linear equations that couple transition probabilities with remaining expectations. Solving the system yields the exact expectation of 6 flips, demonstrating tight coordination between probabilistic and algebraic reasoning. In Figure \ref{fig:app_qua}(e), although GRPO correctly enumerates 74 favorable pairs, it proceeds to ``simplify'' the fraction and erroneously converts 105 into 52.5, corrupting the final answer. GCPO completes the combinatorial count and then explicitly checks the greatest common divisor, confirming that 74/105 is already in the lowest terms and preserving numerical integrity.
Also, in Figure \ref{fig:app_qua}(f), GRPO considers only the modulus-8 requirement that n be odd, ignoring the modulus-3 condition, and consequently overestimates the solution set. GCPO decomposes the problem into two modular layers, i.e., oddness (mod 8) and non-multiplicity by 3 (mod 3), and performs inclusion exclusion counting to arrive at the exact total of 334 integers. Therefore, GCPO systematically integrates local conditions and cross-component interactions, ensuring logical consistency and robust error correction throughout the reasoning chain.

\end{document}